\documentclass{article}

\usepackage{svg}

\usepackage[final, nonatbib]{neurips_2024}
\usepackage[utf8]{inputenc} %
\usepackage[T1]{fontenc}    %
\usepackage{hyperref}       %
\usepackage{url}            %
\usepackage{booktabs}       %
\usepackage{amsfonts}       %
\usepackage{nicefrac}       %
\usepackage{microtype}      %
\usepackage{xcolor}         %
\usepackage{amsmath,amsthm,amssymb}
\usepackage{algorithm}
\usepackage{algorithmic}

\usepackage{subcaption}
\usepackage{multirow}

\usepackage{xspace}
\usepackage{wrapfig}

\usepackage{wrapfig}
\usepackage{algorithm}
\usepackage{algorithmic}

\usepackage{listings}
\usepackage{xcolor}
\usepackage{graphicx}
\usepackage[utf8]{inputenc} 
\usepackage[T1]{fontenc}    
\usepackage{hyperref}       
\usepackage{url}            
\usepackage{booktabs}       
\usepackage{amsfonts}       
\usepackage{nicefrac}       
\usepackage{microtype}      
\usepackage{xcolor}         
\usepackage{enumitem}

\newcommand{\vQ}{\mathbf{Q}}
\newcommand{\vK}{\mathbf{K}}
\newcommand{\vV}{\mathbf{V}}

\newcommand{\vO}{\mathbf{O}}

\newcommand{\sysname}{\textsc{Mini-Sequence Transformer}\xspace}

\newcommand{\shortname}{\textsc{MsT}\xspace}

\newtheorem{theorem}{Theorem}

\usepackage{threeparttable}

\title{\sysname: Optimizing Intermediate Memory for Long Sequences Training}

\author{%
  Cheng Luo\\
  California Institute of Technology \\
  \texttt{chengluo@caltech.edu}\\
  \And
  Jiawei Zhao \\
  Meta FAIR \\
  \texttt{jwzhao@meta.com}\\
  \And
  Zhuoming Chen \\
  Carnegie Mellon University \\
  \texttt{zhuominc@andrew.cmu.edu}\\
  \And
  Beidi Chen \\
  Carnegie Mellon University \\
  \texttt{beidic@andrew.cmu.edu}\\
  \And
  Anima Anandkumar \\
  California Institute of Technology \\
  \texttt{anima@caltech.edu}\\
}

\begin{document}

\maketitle


\begin{abstract}
We introduce \sysname (\shortname), a simple and effective methodology for highly efficient and accurate LLM training with extremely long sequences. \shortname partitions input sequences and iteratively processes mini-sequences to reduce intermediate memory usage. Integrated with activation recomputation, it enables significant memory savings in both forward and backward passes. In experiments with the Llama3-8B model, with \shortname, we measure no degradation in throughput or convergence even with 12x longer sequences than standard implementations. \shortname is fully general, implementation-agnostic, and requires minimal code changes to integrate with existing LLM training frameworks. Integrated with the huggingface library, \shortname successfully extends the 
maximum context length of Qwen, Mistral, and Gemma-2 by 12-24x. 
\end{abstract}

\section{Introduction}
\label{Introduction}


The development of Transformer \cite{vaswani2017attention} has been a remarkable journey, with each iteration pushing the boundaries of what is possible regarding model size, performance, and efficiency. One of the critical challenges in this journey has been managing the memory requirements of these models, particularly during training. As Transformers have significantly grown in size\cite{chen2023long} and complexity \cite{raffel2020exploring}, the memory demand has increased exponentially, necessitating innovative solutions to optimize memory usage while maintaining performance.

A significant milestone in this journey was the introduction of multi-query attention \cite{shazeer2019fast}. This technique dramatically reduced the size of the KV-cache during inference, which uses multiple query heads but single key and value heads. The idea was first adopted in the large-scale training of PaLM \cite{chowdhery2023palm}, then adopted and empirically tested in LLaMA \cite{touvron2023llama}. As the field progressed, multi-query attention evolved into grouped query attention (GQA) \cite{ainslie2023gqa}, which relaxes the single key and value head restriction to multiple heads, and each head is coupled with a group of queries. It significantly improves the quality and is adopted by Llama2-70B \cite{touvron2023llama} and Mistral-7B \cite{jiang2023mistral}.

To further improve model quality, Llama3 \cite{llama3} introduced a tokenizer with a vocabulary of 128K tokens, enabling more efficient language encoding than Llama2's 32K vocabulary. Additionally, Llama3 increased its MLP intermediate size from 11k to 14k. These changes reflect a trend toward more extensive vocabulary and intermediate sizes for better quality. Meanwhile,  Llama3 maintains its hidden size of 4k for inference efficiency. This trend is also reflected in the Microsoft development of Phi-3 \cite{abdin2024phi} compared with Phi-2 \cite{javaheripi2023phi}.

These advancements have also brought about new memory challenges, particularly in the intermediate value of linear layers of multilayer perception (MLP) and language modeling head (LM-Head). The substantial increase in intermediate variables, which can be nearly ten times larger than the input variables, has severely limited the network's ability to expand sequence length and batch size. This limitation has made it difficult to train large models without restricting sequence length to 8K or relying on gradient accumulation or distributed systems to expand batch size.

\begin{figure}[t]
\begin{center}
  \includegraphics[scale=0.5]{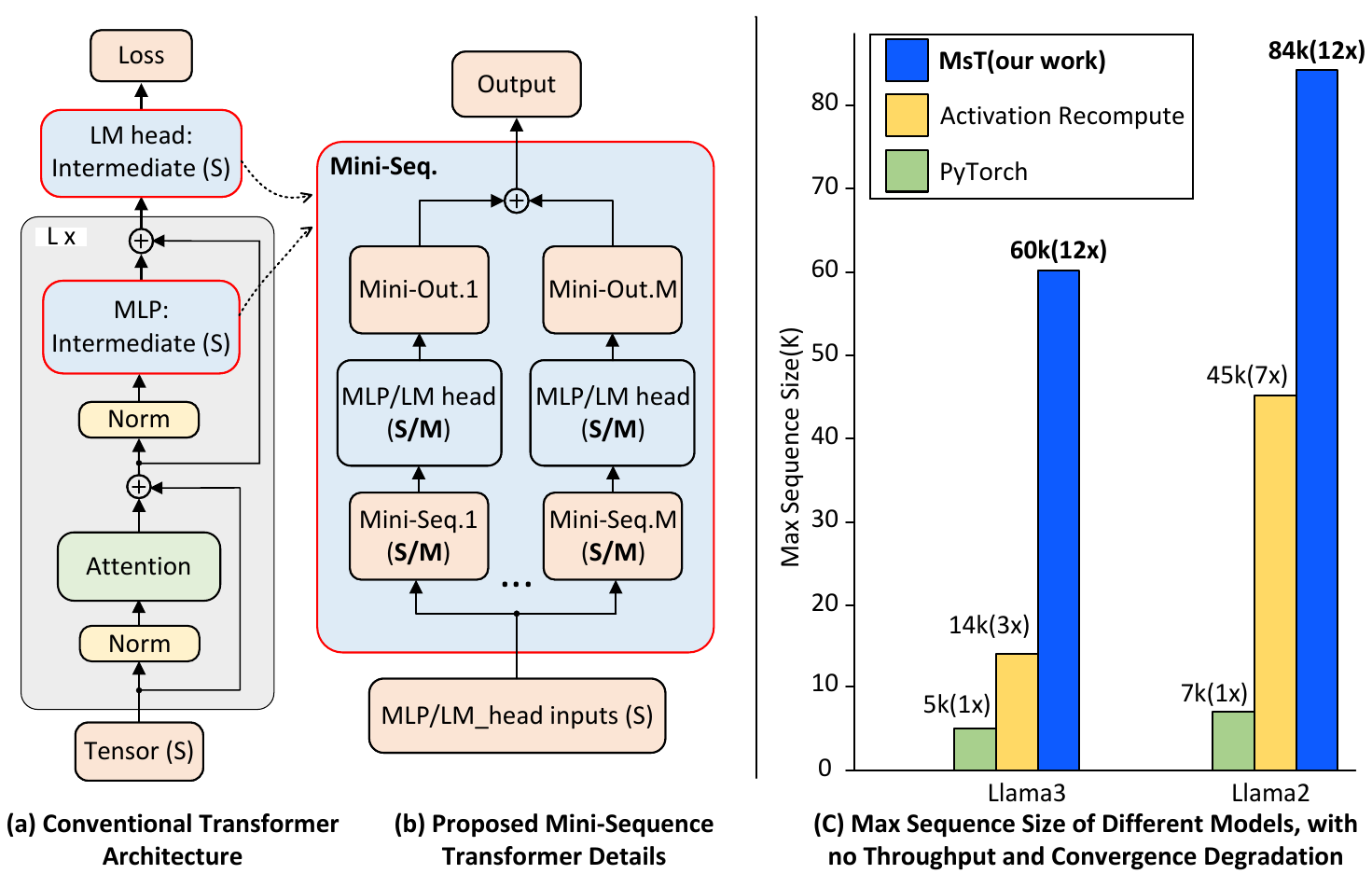}
  
  \caption{(a) Standard Transformer architecture. MLP's and LM-Head's activation sequence length is annotated with $S$. (b) \sysname is used to replace MLP blocks and LM-Head block, which splits the input sequence $S$ into $M$ mini-sequences with sequence length $S/M$, where $M=2$ on this figure. (c) Max sequence size for training Llama2/Llama3 on A100-80GB GPU, with no degradation of throughput or convergence using our approach.}
  \label{fig:transformer-general}
    \vspace{-2em}
\end{center}
\end{figure}

{\bf{Our Approach:}} Recognizing these challenges, we introduce \sysname (\shortname), a simple and effective methodology for enabling highly efficient and highly accurate LLM training with extremely long sequence lengths by reducing intermediate memory overhead.  \shortname introduces a per-layer mini-sequence where the input partitions work for each MLP and LM-Head block. \shortname partitions individual samples along the sequence dimension and iteratively processes each mini-sequence, combining all mini-sequence results to recover full-sequence outputs for these blocks. Our work also adopts activation recomputation \cite{chen2016training}. 
We find no degradation in throughput or convergence even with sequences up to $12\times$ compared to a standard implementation of Llama3-8B, as shown in Figure \ref{fig:transformer-general}(c). 
 

To summarize, we make the following contributions to advance the long-sequence training:
\begin{itemize}[leftmargin=*]
\item  \shortname trains $12-24 \times$ longer sequence lengths than existing systems on a single A100 GPU with no degradation in throughput and convergence of training.
\item  Fully general and implementation agnostic:  \shortname supports most parameter-efficient training as it works independently with attention layers.
\item  Support for large-scale distributed training: \shortname works together with DeepSpeed-Ulysses \cite{jacobs2023deepspeed} to support linear scaling sequence length by the number of GPUs.
\item  Easy-to-use and portable, requiring minimal code changes to the existing training frameworks like Huggingface \cite{jain2022hugging}. The details can be referred to Appendix \ref{Intergration}.
\end{itemize}
In subsequent sections, we provide background and related work, a detailed discussion of  \sysname (\shortname) design, Hardware-efficient analysis, experimental evaluation, and comparison with existing work. This work is open-source under an MIT license on \href{https://github.com/wdlctc/mini-s}{https://github.com/wdlctc/mini-s}.

\section{Background and Related Work}
\label{gen_inst}

This section briefly overviews the performance characteristics of long sequence transformers on modern hardware(e.g., GPUs). We also describe some backgrounds of mini-batch training and activation recomputation, which inspire our work.

\subsection{Transformer Architecture}

Figure \ref{fig:transformer-general}(a) is a sketch of the building blocks of a typical Transformer architecture \cite{vaswani2017attention}. It consists of input sequences $S$ sent into $L$ repeated block with attention and MLP, then computed output loss with LM-Head block. The inputs and outputs of each block are typically a 3D tensor of size $(B, S, d)$ where $B$ is micro batch size, $S$ is sequence length, and $d$ is hidden dimension. The intermediate value includes the $Q, K, V$ tensors of size $(B, S, d)$  within the attention block, the $I$ tensor of size $(B, S, I)$  within the MLP block, and the logits tensor of size  $(B, S, V)$ of within LM-Head block. Here, $I$ represents the intermediate size of MLP, and $V$ represents the vocabulary size.

\subsection{Hardware Performance of Long Sequence Training}


\textbf{Memory Hierarchy.} GPUs have a memory hierarchy with larger but slower global GPU memory (high bandwidth memory; HBM) and smaller but faster-shared memory (SRAM). Transformers' high memory demand originates from the quadratic complexity of self-attention operations, where the memory needed to store attention scores for each token increases quadratically as the sequence length grows. This dramatic increase in memory demand can quickly overwhelm the capacity of the HBM, leading to OOM issues. Flashattention \cite{dao2022flashattention} uses kernel fusion to effectively mitigate the memory overheads associated with the quadratic growth in sequence length, and Xformer \cite{rabe2021self} deploys optimized memory access patterns that achieve linear memory scaling. Our work is partly inspired by memory optimization technologies, where our optimization targets are MLP and LM-Head.

\textbf{Occupancy.} GPUs have many threads executed in parallel; threads are grouped into thread blocks, which execute on streaming multiprocessors (SMs). Modern hardware has specialized units like tensor cores on NVIDIA GPU to accelerate mammals. In long sequence training scenarios where the sequence size tends to be long (>10k), parallelizing over the sequence dimension usually enables high GPU occupancy.

\textbf{Performance characteristics.} GPU operators can be classified as either compute-bound or memory-bound, which is determined by the time spent in arithmetic operations and the time spent accessing HBM. Typical self-attention with long sequence, MLP with the long intermediate size is a compute-bound operator because their core operators are matrix-multiply with a large inner
dimension of sequence length. Then, cross-entropy with reduction is memory-bound.

\subsection{Mini-Batch Training }
Our work is inspired by Mini-Batch Training algorithms, also known as gradient accumulation. Mini-batch training algorithms  \cite{hermans2017accumulated, ott2018scaling} can support large batch size by processing the training batch in smaller mini-batches, which allows the model to be trained on a subset of the data at a time, accumulating gradients over several mini-batch and only updating the parameter with accumulated gradient. 
This reduces the memory requirements compared to batch gradient descent \cite{khirirat2017mini}, which enables training bigger batch sizes than GPU memory constrain. We are inspired by the idea and adapt it to train long sequences instead of large batch sizes.

\subsection{Activation Recomputation }
Activation recomputation \cite{chen2016training}, also known as gradient checkpointing, is a memory-saving technique for training large neural networks. This method trades computation for memory by discarding intermediate activations during the forward pass and recomputing them as needed during the backward pass. In standard training, all activations must be stored to compute gradients, which can lead to significant memory usage for large models or long sequences. 
Activation recomputation is orthogonal with our \shortname, and we integrate this method for better optimizing intermediate value. We analyze the memory efficiency of activation recomputation and its integration with \shortname on Sec \ref{thm:memory}.

\section{\sysname (\shortname): Algorithm, Analysis, and Distributed Extensions}

We present our \sysname (\shortname) mechanism to partition the input sequence into $M$ mini-sequences. We show how to compute the exact transformer block by gradient accumulation during the backward pass. Then, we analyze its memory efficiency and IO complexity, showing that our method is memory-efficient and throughput-equalized compared to the standard transformer. Based on the analysis, we found the optimal implementation of \shortname by selecting the best hyperparameters. We further show how \shortname can work on distributed settings by integrating with DeepSpeed \cite{jacobs2023deepspeed}.

We focus here on the forward pass for ease of exposition; Appendix \ref{sec: Algorithm Details} contains details for the backward. 

\subsection{Algorithms: Optimizing Intermediate Memory With Mini-Sequence Processing }

Our idea arises from the observation of large intermediate values from transformer blocks. Given the inputs  $X \in \mathbb{R}^{N \times d}$ in HBM, attention blocks and MLP blocks compute the output $O \in \mathbb{R}^{N \times d}$ and LM-head block computes the output $loss \in \mathbb{R}^{1}$, $N$ equals to sequence size $S$ here. We observe that the intermediate values are always larger than the input $X$ and output $O$, $loss$, illustrated in Table \ref{table:intermediate}. Attention has intermediate values $\vQ, \vK, \vV \in \mathbb{R}^{N \times d}$, which is $(1 + 2\times d)/G$ larger than input size, where  $(1 + 2\times d/G=1.5)$ in Llama3 setting. $G$ refers to the number of grouped query attention (GQA). MLP has intermediate value $I_{up}, I_{gate} \in \mathbb{R}^{N \times I}$, where $2\times I/d = 7$ in Llama3 setting. LM-Head has $logits \in \mathbb{R}^{V \times d}$, where $V/d = 32$ in Llama3 setting. The detail setting of Llama3-8B is listed in Appendix \ref{sec: Model Architecture Comparison}

\begin{table}[h]
\vspace{-3mm}
  \small
  \centering
  \caption{\label{table:intermediate}Intermediate value size analysis for transformer blocks}
  \setlength{\tabcolsep}{5pt}
  \begin{threeparttable}
  {
    \begin{tabular}{@{}c|ccc@{}}
      Transformer Blocks & Input/Output Size &\multicolumn{1}{c}{Peak Intermediate Value Size}&\multicolumn{1}{c}{ Intermediate/Input Ratio \tnote{1}} \\
    \hline
      Attention & $(B, S, d) /  (B, S, d)$ & $(B, S, d) + 2\times(B, S, d/G)$ & $(1 + 2\times d/G) \approx 1.5 $  \\
      MLP &  $(B, S, d) /  (B, S, d)$ & $2 \times (B, S, I)$ & $(2 \times I) / d \approx 7$ \\
      LM-Head &  $(B, S, d) /  1$ & $(B, S, V)$ & $V/d \approx 32$ \\
    \end{tabular}
    \footnotesize
    \begin{tablenotes}
    \item[1] The ratio in Llama3 setting.
    \end{tablenotes}
  }
  \end{threeparttable}
\vspace{-1mm}
\end{table}

As flash attention and group query attention have minimized the intermediate value of attention, we put our focus on the MLP block and LM-Head block. Therefore, our implementation of \shortname is general enough to work with any attention: self-attention \cite{vaswani2017attention}, cross-attention \cite{bahdanau2014neural}, causal attention \cite{radford2018improving}, their sparse counterparts \cite{child2019generating, zaheer2020big, roy2021efficient}, and their various optimized kernels such as different versions of FlashAttention \cite{dao2022flashattention, dao2023flashattention}. Our implementation adopts FlashAttention2 \cite{dao2023flashattention} for the experiments.

\paragraph{Input Partition.} We apply the mini-sequence technique to overcome the technical challenge of large intermediate values occupying HBM memory. We describe this in Algorithms \ref{alg:stream_MLP}, and \ref{alg:stream_o}, which represent MLP blocks and LM-Head from Llama serials. Their MLP block consists of three linear layers and SiLU function \cite{ramachandran2017searching}, and their LM-Head block consists of one linear layer and CrossEntropyLoss function\cite{rubinstein1999cross}. The corresponding backward implementations can be referred to in Appendix \ref{sec: Algorithm Details} for more details. The main idea is to partition the input $X$ into mini-sequence $X_i$ as Algorithm \ref{alg:stream_MLP} line 1 and Algorithm \ref{alg:stream_o} line 1, then compute the output with respect to those mini-sequences. We get the exact same result as standard implementation by contacting all mini-sequence outputs.

\paragraph{Gradient Accumulation.} One of our goals is to reduce intermediate values for backward passes. The backward pass typically requires the matrices $X \in \mathbb{R}^{N \times d}$, $I \in \mathbb{R}^{N \times I}$, $logits \in \mathbb{R}^{N \times V}$ to compute the gradients with respect to weights. However, by input partition the $X \in \mathbb{R}^{N_m \times d}$, we can reduce the intermediate value as $I \in \mathbb{R}^{N_m \times I}$, $logits \in \mathbb{R}^{N_m \times V}$ by $M \times$ in the backward pass in HBM. With gradient accumulation for all mini-sequences, all gradients are generated in the same way as standard implementation by introducing more memory loading time. However, as MLP is the standard computation-bound operator and LM-Head occupies only a small amount of total training time, \shortname would not affect the whole training speed with a significant reduction in memory overhead. 

\vspace{-0.5em}
\begin{algorithm}[H]
  \caption{\small\label{alg:stream_MLP}Mini-Sequence MLP}
  \begin{algorithmic}[1]
    
    \REQUIRE Matrices $X \in \mathbb{R}^{N \times d}$, MLP block, $W_{down}, \in \mathbb{R}^{I \times d}, $  Weights of three linear layers $W_{gate}, W_{up} \in \mathbb{R}^{d \times I}$, $W_{down} \in \mathbb{R}^{I \times d}$

    \STATE Partition matrices $X$ into $M$ blocks $X_1, \dots, X_m$ of size $N_m \times d$, where $N_m = N/M$
    
    \FOR{$1 \le i \le M$} 
    \label{alg:stream_attn_outer_loop}
    \STATE Compute  $\vO_{i}' = MLP(X_i, W_{gate}, W_{up}, W_{down}) $, $\vO_{i} \in \mathbb{R}^{N_m \times d}$
    
    \ENDFOR
    \STATE Contact $\vO = \{\vO_i', \dots, \vO_m'\} \in \mathbb{R}^{N \times d}$
    \STATE Return $\vO$.
  \end{algorithmic}
\end{algorithm}
\vspace{-1em}

\label{headings}




\vspace{-0.5em}
\begin{algorithm}[H]
  \caption{\small\label{alg:stream_o}Mini-Sequence LM-Head}
  \begin{algorithmic}[1]
    \REQUIRE Matrices $X \in \mathbb{R}^{N \times d}$, Labels $L \in  \mathbb{R}^{N}$, Weights $W_{out} \in \mathbb{R}^{d \times V}$

    \STATE Partition matrices $X$ into $M$ blocks $X_1, \dots, X_m$ of size $N_m \times d$, where $N_m = N/M$
    \STATE Partition labels $L$ into $M$ sub-label, $L_1, \dots, L_m$ of size $N_m$, where $N_m = N/M$
    
    \FOR{$1 \le i \le M$} 
    \STATE Compute  $logits_{i} = X_i W_{out} $, $logits_{i} \in \mathbb{R}^{N_m \times V}$
    \STATE Compute if  $ (i-1)*N_m \leq L_{i} \le (i-1)*N_m$, $ L_{i}=L_{i}$ else $L_{i} = -100 $
    \STATE Compute  $loss_{i} = crossentropyloss(logits_{i}, L\_) $
    
    \ENDFOR
    \STATE Compute $ loss = \sum_1^M loss_i / M$
    \STATE Return $loss$.
  \end{algorithmic}
\end{algorithm}
\vspace{-2em}




\subsection{Analysis: Memory Efficiency  of \sysname (\shortname)}
\label{thm:memory}
We analyze the memory efficiency of \shortname. MST can reduce intermediate value by $M \times$ while maintaining the same throughput performance.

\begin{theorem}\label{thm:original memory}
  Let $S$ be the sequence length, $W_{mem}$ be the weight memory occupation, including weights, gradient, and optimizer. $A_{mem}$ be the activation memory occupation per sequence, $I_{mem}$ be the intermediate memory occupation per sequence. The peak memory of the standard transformer is achieved by $ M = W_{mem} + S \times (I_{mem} + L \times A_{mem})$. Note that $ L \times A_{mem} >> I_{mem}$ for standard transformer, as $A_{mem}$ lasts for all $L$ layers, but $I_{mem}$ only lasts for one layer.
\end{theorem}

\begin{theorem}\label{thm:activation recomputation}
  With OpenAI's activation recomputation\cite{OpenAIgradient}, the  $ L \times A_{mem}$ could be reduced to $sqrt(L) \times A_{mem}$. Therefore the peak memory is reduced to $ M = W_{mem} + S \times (I_{mem} + sqrt(L) \times A_{mem})$. For models with a large vocabulary and MLP intermediate, $sqrt(L) \times A_{mem} < I_{mem}$.
\end{theorem}

\begin{theorem}\label{thm:MST memory}
  MST can reduce intermediate value by $M \times$, so 
  the memory occupation becomes $ M = W_{mem} + S \times (I_{mem} / M + sqrt(L) \ times A_{mem})$. For GPU with maximum memory $M_{max}$, the maximum sequences length is contained by $S_{max} = \frac{(M_{max} - W_{mem})}{(I_{mem} / M + sqrt(L) \times A_{mem})}$.  This sequence length would be much longer than the standard implementation with $S_{max} = \frac{(M_{max} - W_{mem})}{(I_{mem}  + L \times A_{mem})}$.
\end{theorem}

\subsection{Analysis: IO Complexity and Memory of \sysname (\shortname)}
\label{Analysis}

We analyze the IO complexity of \shortname, compared with consistent compute complexity, which can affect its compute-bound or memory-bound performance characteristics.

\begin{theorem}\label{thm:computation_complexity}
  Let $S$ be the sequence length, $d$ be the hidden dimension, $I$ be the intermediate size, and $V$ be the voice size.
  Standard MLP returns $O = act ((X W_{gate}) * (X_i W_{up})) * W_{down} $ with $O(SdI)$ FLOPS and \shortname MLP returns $O(SdI / M * M) = O(SdI)$ FLOPS.
  Standard LM-Loss returns $loss = crossentropyloss(XW, L) $ with $O(SdV + SV)$ FLOPS, and \shortname LM-Loss returns $O((SdV + SV) / M * M) = O(SdV + SV)$ FLOPS. 
\end{theorem}

\begin{theorem}\label{thm:io_complexity}
  Standard MLP requires $\Theta(Sd + SI + dI)$ HBM
  accesses, while \shortname (\ref{alg:stream_MLP}) requires $\Theta(Sd + SI + dIM)$ HBM accesses.
  Standard LM-Head requires $\Theta(Sd + SV + dV)$ HBM
  accesses, while \shortname (\ref{alg:stream_o}) requires
  $\Theta(Sd + SV + dVM)$ HBM accesses.
\end{theorem}

For Llama3 values of $d$ (4096), $I$ (14336) and $V$ (128256), $SI$, $Sv$ is many time larger than $Sd$. For long sequence cases, the compute complexity and IO complexity are dominated by $SI$ and $SV$, where \shortname is close to standard implementation. 
However, for small sequence cases where $S << d$, the compute complexity and IO complexity are dominated by $dI$ and $dV$ while \shortname needs $dIM$ and $dVM$. Therefore, \shortname would cause throughput downgrades for small sequence lengths.


\subsection{Chunk-based \sysname (\shortname)}
\label{Chunk}

We present an optimized implementation of chunk-based \shortname designed to mitigate throughput reductions when training with small sequence data. The fundamental approach involves partitioning sequences $S$ into equally sized chunks of size $C$ (when possible), resulting in $M = S/C$ mini-sequences.

Our IO complexity analysis indicates that the number of mini-sequences $M$ influences the HBM accesses as $\Theta(Sd + SI + dIM)$ and $\Theta(Sd + SV + dVM)$. However, the HBM accesses remain stable at $\Theta(SI)$ and $\Theta(SV)$ provided that $dIM \leq SI$ and $dVM \leq SV$. It means $d \leq S/M$.

Therefore, by setting the chunk size to $C = S/M \geq d$, MST avoids throughput downgrades for small sequences. Intuitively, when the sequence size is smaller than the chunk size, MST does not split the input, thereby preventing any performance loss.



We apply chunk-based \shortname exclusively to MLP blocks by setting a constant chunk size $C$ equal to the hidden dimension, $C = d$. For LM-head blocks, we maintain a constant mini-sequence size of $M = V/d$, as these blocks contribute minimally to the overall training time of transformers.





\begin{wrapfigure}{R}{0.25\textwidth}
  \vspace{-2em}
  \centering
  \includegraphics[scale=0.55]{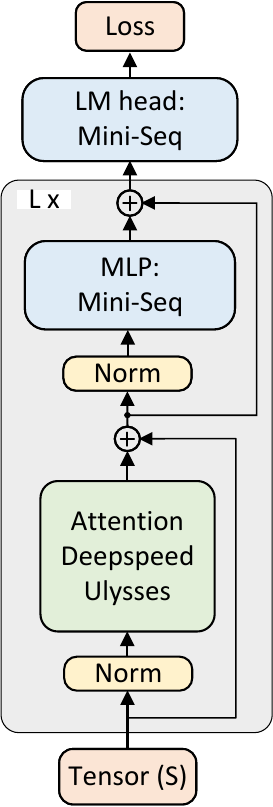} 
  \caption{ Distributed \sysname. }\label{fig:narrowtransformer_deepspeed}
  \vspace{-3em}
  
\end{wrapfigure}

\subsection{Extension: Distributed \sysname (\shortname)}
We extend \sysname (\shortname) to the distributed setting: we propose \shortname+ SP, which can effectively scale the transformer using sequence parallelism(SP). In SP, the input tensor of each Transformer layer is divided along the sequence dimension, allowing for parallel computation across multiple GPUs. This segmentation, in conjunction with activation recomputation, results in a substantial reduction in activation memory requirements. It is worth noting that our proposed approach is orthogonal to most sequence parallelism, such as Megatron-LM \cite{korthikanti2023reducing}, Deepspeed-Ulysses \cite{jacobs2023deepspeed}, Sequence parallelism \cite{li2021sequence}, and Ring Attention \cite{liu2023ring}. Here, we take Deepspeed-Ulysses as an example of how they work together.

Figure \ref{fig:narrowtransformer_deepspeed} shows the design of extending \shortname with DeepSpeed-Ulysses. As with the transformers architecture, the design consists of an attention block with  DeepSpeed-Ulysses, MLP, and LM-Head with \shortname's mini-sequence technology. The design consists
of input sequences $S$ partitioned across available devices and mini-sequences. Each attention block  Matrices $\vQ, \vK, \vV $ are communicated through all-to-all collectives before and after the attention computation. The remaining modules of MLP and LM-Head use the sequence parallel and mini-sequence together. As DeepSpeed-Ulysses's main change is working on attention block and \shortname is working on MLP and LM-Head, it is straightforward to make them work together to scale sequence length. 


\section{Experiment}
\label{others}

We evaluate the impact of using chunk-based \sysname(\shortname) on Llama3\cite{llama3}, a state-of-the-art model for many NLP tasks. We also evaluate Qwen \cite{bai2023qwen}, Mistral \cite{jiang2023mistral}, and Gemma-2 \cite{team2024gemma} for context length improvements. We validate our claims about scaling sequence length, reporting training time, and memory overhead. Distributed Extension results can be found in appendix \ref{distributed experiment}, which confirms that the sequence length of \shortname can scale linearly with the number of GPUs.

\begin{itemize}[leftmargin=*]
    \item \textbf{Maximun Sequence Length.} \shortname can train Llama3-8B with context length 60k and Llama3-7B with context length 84k on a single A100 GPU, outperforming the standard implementation by $12\times$. Also, it achieves $12-24\times$ than the standard implementation of Qwen, Mistral, and Gemma-2.
    \item \textbf{Training throughput.} \shortname maintains the same training throughput compared with standard long-sequence training. Moreover, the throughput can be slightly improved with a large batch size supported by \shortname.
\end{itemize}

\subsection{Longer Sequence Length with \sysname (\shortname)}

\paragraph{Llama3 and Llama2.} We train a Llama3-8B\cite{llama3} \shortname and Llama2 models\cite{radford2019language}  \shortname  by exploring the sequence length on a single A100 GPU with lossless training strategies, such as activation recomputation, fusing the backward operation with the optimizer update \cite{lv2023adalomo} and \shortname. 
Table \ref{table:llama3_s} compares our maximum sequence and training time to the PyTorch standard implementation and Huggingface PEFT with activation recomputation. Our implementation trains $4\times$ longer sequence LLAMA-3 compared with activation recomputation and $12\times$ longer sequence compared with standard implementation. Also, our implementation trains $1.8\times$ longer sequence compared with activation recomputation and $12\times$ longer sequence compared with standard implementation.


\begin{table}[h]
  \small
  \centering
  \vspace{-1em}
  \caption{\label{table:llama3_s}Maximum sequence length of Llama3-8B and Llama2-7B.
}
  {
    \begin{tabular}{@{}c|c@{}}
      Llama3-8B-hf Implementation & Maximum Sequence Length (K)  \\ \hline
      Llama3-8B-hf vanilla  & 5 \\
      Llama3-8B-hf activation recomputation  & 14  \\
      Llama3-8B-hf \shortname  & 60  \\
       \hline
      Llama2-7B-hf vanilla  & 7 \\
      Llama2-7B-hf activation recomputation  & 45  \\
      Llama2-7B-hf \shortname  & 84  \\
       \hline
    \end{tabular}
  }
  \vspace{-1em}
\end{table}



\paragraph{Qwen, Mistral, and Gemma-2.} We've extended our evaluation to include Mistral-7B, Qwen2-7B, and Gemma-2-9B, demonstrating significant increases in maximum sequence length ($12\times$ for Mistral-7B, $18\times$ for Qwen2-7B, $24\times$ for Gemma-2-9B) across these architectures. Among these models, \shortname provide best sequence extension for Gemma-2 of $24\times$. The critical observation here is that gemma-2 uses the largest vocal size (256k) than Mistral-7B (32k) and Qwen2(152k).

\begin{table}[h]
  \small
  \centering
  \vspace{-1em}
  \caption{\label{table:model_comparison}Maximum sequence length of various models.}
  {
    \begin{tabular}{@{}c|c@{}}
      Model Implementations & Maximum Sequence Length (K)  \\ \hline
      Mistral-7B vanilla & 5 \\
      Mistral-7B activation recomputation & 42 \\
      Mistral-7B MST & 70 \\
      \hline
      Qwen2-7B vanilla & 4 \\
      Qwen2-7B activation recomputation & 13 \\
      Qwen2-7B MST & 74 \\
      \hline
      gemma-2-9b vanilla & 1.5 \\
      gemma-2-9b activation recomputation & 5 \\
      gemma-2-9b MST & 36 \\
      \hline
    \end{tabular}
  }
  \vspace{-1em}
\end{table}

\paragraph{Combination with gradient accumulation.} Gradient Accumulation has been used during training Llama2 and Llama3, which helps them train larger batch sizes given limited available GPU memory. However, in Gradient Accumulation, instead of updating the model parameters after processing each batch of training data, the gradients are accumulated over multiple batches before updating. This means that the memory usage for gradients would occupy the memory used for activation. Therefore, using gradient accumulation during training would constrain the maximum sequence size. 

Table \ref{table:llama3_gradientaccumulation} summarizes the maximum sequence length with gradient accumulation. The activation recomputation technology can train up to 8K sequences. Then \shortname can train up to 30k sequence length, which is 
$4\times$ longer sequence length than activation recomputation, and $21\times$ longer than vanilla.  For Llama2-7B, \shortname can also train up to 55k sequence length.

\begin{table}[h]
  \small
  \centering
  \vspace{-1em}
  \caption{\label{table:llama3_gradientaccumulation}Maximum sequence length training with gradient accumulation.
}
  {
    \begin{tabular}{@{}c|c@{}}
      Model Implementation with gradient accumulation & Maximum Sequence Length (K)  \\ \hline
      Llama3-8B-hf vanilla & 1.5  \\
      Llama3-8B-hf Activation Recomputation  & 8  \\
      Llama3-8B-hf \shortname  & 32  \\
       \hline
      Llama2-7B-hf vanilla & 4  \\
      Llama2-7B-hf activation recomputation  & 38  \\
      Llama2-7B-hf \shortname  & 55  \\
      \hline
    \end{tabular}
  }
  \vspace{-1em}
\end{table}

\paragraph{Comparison and Combination with Lossy Methods.}
We've comprehensively compared \shortname with quantization methods and the combinations between \shortname and quantization on Table \ref{table:llama3_sequence_lengths}. All lossy methods are HuggingFace official implementations. This comparison demonstrates \shortname's superiority in enabling longer sequences for Llama3 training on a single A100 GPU. \shortname alone (60K tokens) outperforms these lossy approaches (4bit 28k). When combined with quantization techniques, \shortname achieves even more impressive results: \shortname + 8-bit reaches 110K tokens (a $22 \times $ improvement over standard 8-bit), while \shortname + 4-bit pushes the boundary to 140K tokens. We did not evaluate the effect of quantization on training loss.

\begin{table}[h]
\small
\centering
\vspace{-1em}
\caption{Maximum sequence length training with lossy method}
\begin{tabular}{@{}c|c@{}}
Llama3 Implementations & Maximum Sequence Length (K) \\ \hline
8-bit                           & 5   \\ 
4-bit                           & 10   \\ \hline
MST                             & 60   \\
MST + 8-bit                     & 110  \\
MST + 4-bit                     & 140   \\ \hline
\end{tabular}
\label{table:llama3_sequence_lengths}

  \vspace{-2em}
\end{table}



  

\subsection{Faster Long Sequence Training with \sysname (\shortname)}
We evaluate the training performance of \shortname on Llama3-8B with 8k sequence and Llama2-7B with 4k sequence using a single A100 80G GPU. Table \ref{table:llama_speed} compares the training time per step and TFLOPS achieved by \shortname with the vanilla PyTorch implementation and activation recomputation technique.

\begin{table}[h]
  \small
  \centering
  \vspace{-1em}
  \caption{\label{table:llama_speed}Training performance using \shortname on single A100 80G GPU.}
  {
    \begin{tabular}{@{}c|c|c|c@{}}
      Model Implementation & Batch Size & Training Time Per Step (s) & TFLOPS  \\
      \hline
      Llama3-8B-hf vanilla & 1  & OOM & OOM \\
      Llama3-8B-hf activation recomputation & 2 & 5.01 & 3271.42  \\
      Llama3-8B-hf \shortname & 2 & 5.13 & 3194.90  \\
      Llama3-8B-hf \shortname & 8 & 19.35 & 3386.13  \\
      \hline
      Llama2-7B-hf vanilla & 1 & 1.24 & 3290.88 \\
      Llama2-7B-hf activation recomputation & 1 & 1.52 & 2684.67 \\
      Llama2-7B-hf \shortname without activation recomputation & 1 & 1.31 & 3115.03 \\
      Llama2-7B-hf activation recomputation & 8  & 8.85 & 3703.48  \\
      Llama2-7B-hf \shortname & 8 & 9.33 & 3511.39  \\
      Llama2-7B-hf \shortname & 16 & 17.92 & 3656.17  \\
      \hline
    \end{tabular}
  }
  \vspace{-1em}
\end{table}

For Llama3-8B, the vanilla implementation runs out of memory (OOM) with a batch size of 1. Activation recomputation allows training with a batch size of 2, achieving 3271.42 TFLOPS and a training time of 5.01 seconds per step. \shortname, with the same batch size of 2, achieves a comparable 3194.90 TFLOPS with a slightly longer training time of 5.13 seconds per step. However, \shortname's memory efficiency allows scaling the batch size to 8, resulting in an improved 3386.13 TFLOPS and a training time of 19.35 seconds per step.

In the case of Llama2-7B, the vanilla implementation can train with a batch size of 1, achieving 3290.88 TFLOPS and a training time of 1.24 seconds per step. For the same batch size, \shortname without activation recomputation achieves 3115.03 TFLOPS with a training time of 1.31 seconds per step, demonstrating a 16\% speedup over activation recomputation (2684.67 TFLOPS) and only a 5\% slowdown compared to vanilla PyTorch.
\shortname further increases the batch size to 16, maintaining a similar 3656.17 TFLOPS with a training time of 17.92 seconds per step.


\subsection{Better Models with Longer Sequences}

\paragraph{Language Modeling with Long Context.} The memory efficiency of \shortname allows us to increase the context length of llama by $4\times$ than activation recomputation. 
Table \ref{table:finetune_llama2} shows that training Llama3-8B with 30K context length achieved a $2.7\times$ improvement in perplexity compared to the 8K baseline. We train a Llama3-8B\cite{llama3} \shortname on the 
LongAlpaca dataset\cite{chen2023longlora}. The training lasts for two epochs and 10k steps for demonstration. For all implementation, we use the AdamW optimizer \cite{loshchilov2017decoupled}. We use a weight decay of 0.001, gradient clipping of 1.0, and a constant learning rate of 1e-4. All batch sizes equal 16, with a gradient accumulation step of 16. The bf16 precision is also deployed.

\begin{table}[h]
  \small
  \centering
  \vspace{-1em}
  \caption{\label{table:finetune_llama2}LLAMA3-8b with \shortname, with $4 times$ larger context length compared to activation recomputation.
}
  {
    \begin{tabular}{@{}c|c|c|c|c@{}}
      Llama3-8B-hf Implementation & Context length & LongAlpaca-12k (ppl) & loss & Training Time \\ \hline
      Activation Recomputation  & 8k & 9.34 & 2.23 & 25.6 hours  \\
      \shortname  & 8k & 7.41 & 2.00 & 26.5 hours  \\
      \shortname  & 16k & 3.53 & 1.26 & 62.5 hours \\
      \shortname  & 30k & 3.45 & 1.23 & 233 hours \\
       \hline
    \end{tabular}
  }
  \vspace{-1em}
\end{table}

\section{Ablation Study: }

\subsection{Memory Optimization of \sysname (\shortname)}

\shortname introduces a series of memory optimizations to reduce the memory overhead of long-sequence training. To understand the effectiveness of \shortname memory optimizations, we perform an ablation study that incrementally turns off these optimizations (mini-sequence, activation recomputation) and measures the memory requirements. We consider three options: vanilla (standard Pytorch with BF16), activation recomputation only, and \shortname with activation recomputation.





\begin{figure*}[t]

\centering
\begin{minipage}[c]{0.48\textwidth}
  \centering
  \includegraphics[width=\linewidth]{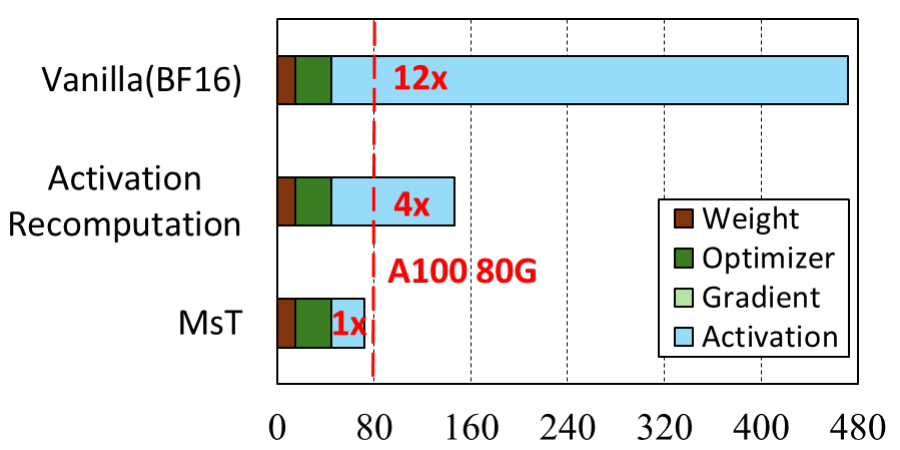}
  (a) Llama3-8B with 20k context.
\end{minipage}
\begin{minipage}[c]{0.48\textwidth}
  \centering
  \includegraphics[width=\linewidth]{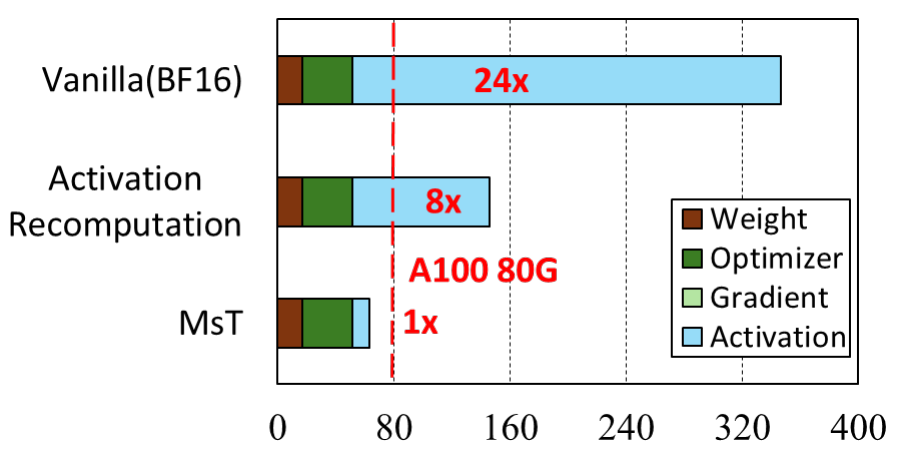}
  (b) Gemma2-9B with 20k context.
\end{minipage}

\caption{Memory consumption of pre-training Llama3-8B and Gemma2-9B
models with a batch size of 1 on a single A100 device, with activation recomputation and \shortname. Note that long-sequence training gradients overlap with activation, so gradients are not shown in bars.}\label{fig:memory}

\vspace{-1.5em}
\end{figure*}

Figure \ref{fig:memory} shows the results. We analyze the peak memory usage of Llama3-8B and Gemma2-9B, with a sequence length of 20k. For sequence length 20k of Llama3-8B and Gemma2-9B, only \shortname can make the model fit into A100 GPU. The rest of the memory consumption is estimated based on its model architectures and theoretical activation amount. For Llama3, activation recomputation can reduce the memory overhead of activation by $3 \times$,  and \shortname can further reduce $4 \times$ memory overhead based on activation recomputation. For Gemma2-9B, \shortname achieves $24 \times$ longer sequence than vanilla and $8 \times$ longer sequence than activation recomputation. This improvement from $12 \times$ to $24 \times$ is due to Gemma2-9B's higher intermediate/input ratio (8 for MLP and 72 for the LM head) compared to Llama3 (7 for MLP and 32 for the LM head, as shown in Table \ref{table:intermediate}).
Further details on the memory ablation study can be found in Appendix \ref{sec:Memory Optimization Detials}.

\subsection{How many mini-sequences are needed during training}

We observe that increasing $M$, the number of mini-sequences, can enhance memory efficiency; however, this enhancement has a certain upper limit. Specifically, increasing $M$ can also affect throughput performance. Appendix \ref{M number} provides details regarding these limitations and their effects.  
This observation allows us to identify the optimal configuration for memory optimization and achieve the best balance between memory performance, consistent with our analysis in Sec \ref{thm:memory} and \ref{Analysis}.



We found that the best balance for memory and throughput is achieved by the optimal values of $C$ for chunk-based MLP $C = d, M = S / d$, where $d$ is the hidden size. For the LM-Head, the original MST is employed for memory saving, and the optimal setting for $M$ is determined by $M = V/d$, specifically 32 for Llama3 and 64 for Gemma-2. This value provides the best memory efficiency.

\section{Limitations and Future Directions}
\label{sec:discussion}

We discuss the limitations and future directions. Related work is also given in Appendix \ref{sec: Related Work}.

\textbf{Compiling to CUDA.} Our current approaches are built on Pytorch implementation. 
This may constrain performance and low-level memory savings. It can be improved by fused kernel and cuda optimization, which can be our next step. 

\textbf{Combination with memory optimization.} Our goal is to increase sequence length while maintaining performance and accuracy. Relaxing these requirements, MsT can be combined with activation offload to extend sequence length as  $S_{max} = \frac{(M_{max} - W_{mem})}{(I_{mem} / M + A_{mem})}$, or with quantization to extend sequence length as  $S_{max} = \frac{bf16}{4bit / 8bit}\frac{(M_{max} - W_{mem})}{(I_{mem} / M + L \times A_{mem})}$. This combination can be explored in future research.


\section*{Acknowledgements}
We thank Vast AI for computational resources renting.

A. Anandkumar is supported by the Bren named chair professorship, Schmidt AI 2050 senior fellowship, ONR (MURI grant N00014-18-12624).
\label{sec:Acknowledgements}

\bibliographystyle{plain}
\bibliography{ref}


\appendix

\section{Related Work}
\label{sec: Related Work}

\paragraph{Long Sequences Model.} 

The ability to train large models with long sequences is becoming increasingly important across various domains, from generative AI to scientific discovery. In generative AI, tasks such as conversational AI, knowledge-rich long document summarization, and video generation demand reasoning over extended contexts in both spatial and temporal dimensions. Multimodal foundation models process speech, images, and waveforms simultaneously, requiring long context reasoning over high-dimensional inputs with lengthy sequences. Similarly, chapter and book-level summarization, estimated to involve tens to hundreds of thousands of words, hold great importance in conversational AI and abstractive summarization tasks \cite{beltagy2020longformer, kryscinski2021booksum, loukas2021edgar} and have demonstrated benefits from long sequence training \cite{xiong2023effective, reid2024gemini, llama3}. 

The emergence of ChatGPT and subsequent open-source and commercial large language models has propelled chat applications to the forefront of modern AI, making them more relevant than ever. Efficiently processing long sequences is vital for supporting more extensive conversation histories in these applications \cite{touvron2023llama}. Long sequence capability is equally important for AI in scientific fields, enabling better understanding and advancements in healthcare \cite{liu2023large}, climate and weather forecasting \cite{nguyen2023climax}, and large-scale molecular simulations \cite{zvyagin2023genslms}.

\paragraph{Lossy Long Sequence Training.} 
One direction is making LLM able to
process arbitrarily long sequences efficiently by sacrificing the perception window of the network. Sliding window attention is introduced \cite{dai2019transformer} to handle infinitely long sequences as input. However,
it disregards information beyond the effective receptive field. Longformer \cite{beltagy2020longformer} extends this idea, which caches on a block-by-block basis, which increases the perception window size, so as TransformerFAM \cite{hwang2024transformerfam}. StreamLLM \cite{xiao2023efficient} is not constrained by a given window but selectively disregards information between the first token and given windows. They struggle to capture long-range dependencies beyond this fixed range, but the approximation quality also seems to degrade at long sequence lengths. \shortname can work directly with them to increase the window size for better quality.

\paragraph{Memory Efficient Training.}
As the demand for long sequence processing continues to grow across various domains, developing efficient methods for training large models with extended context becomes increasingly essential for advancing the state of the art in AI. 
Parameter-efficient fine-tuning (PEFT) methods aim to reduce the memory footprint and computational related to parameters and gradients. Adafactor \cite{shazeer2018adafactor} achieves sub-linear memory cost by factorizing the second-order statistics using a row-column outer product. Low-Rank Adaptation (LoRA) \cite{hu2021lora} reduces the memory footprint of pre-trained models using low-rank adaptors with a low-rank weight adaptor for each layer. Several variants of LoRA have been proposed to enhance its performance. \cite{sheng2023s, zhao2024galore}.
Quantization is another widely used technique in PEFT to reduce the memory cost of optimizer states  \cite{dettmers2024qlora}.

Other techniques focus on reducing the memory footprint and computational related to activation. Gradient accumulation is a method that allows for training larger batch sizes by accumulating gradients over multiple mini-batches before performing an optimization step \cite{ott2018scaling}. This approach enables training with larger effective batch sizes while maintaining a smaller physical batch size, reducing activation memory requirements. A similar work of activation offloading \cite{rajbhandari2021zero} moves the checkpointed activations to the CPU asynchronously and prefetches the offloaded activations back from the CPU during backward.
There are related works in sparse Transformers, mainly focusing on full-attention approximation, such as sparse attention \cite{child2019generating, zaheer2020big}.
Recent works have also focused on single-GPU memory and compute-efficient attention. A popular example in this category is Flash attention \cite{dao2022flashattention}, which leverages known techniques such as tiling and recomputation for computing and memory efficiency. Also, some work put their interest in cross-entropy. FlashCE \cite{apsod2023flashce}  optimizes cross-entropy by leveraging sparse data structures and CUDA optimizations to enhance speed and memory efficiency. Efficient cross-entropy \cite{malek2023efficient_cross_entropy} introduces a memory-efficient variant of cross-entropy loss to reduce activation memory by storing only essential computations.
These works are orthogonal to our work and can be leveraged accordingly to further improve the efficiency of Transformer-based models.


\paragraph{Distributed Training.} Distributed training techniques have become essential for training large language models (LLMs) due to their immense computational and memory requirements. By splitting workloads across multiple GPUs, these methods help alleviate memory bottlenecks and enable the training of models that would otherwise be infeasible on a single device. Data parallelism \cite{krizhevsky2017imagenet} replicates the model on multiple devices, processing different data batches in parallel and synchronizing gradients across GPUs. Tensor parallelism \cite{shoeybi2019megatron} divides individual layers of the model across GPUs, enabling more efficient memory usage for extremely large models.
Another prominent method is fully sharded data parallelism (FSDP) \cite{zhao2023pytorch}, which extends the data parallel approach by sharding both model parameters and optimizer states across devices, thus further reducing memory overhead. Sequence parallelism \cite{korthikanti2023reducing,jacobs2023deepspeed,li2021sequence,liu2023ring} specializes in optimizing memory usage for transformer-based models by partitioning sequences across GPUs and reducing activation memory.
In addition to these, pipeline parallelism \cite{huang2019gpipe} splits the model into segments, with each segment assigned to a different GPU, and processes data in a pipeline fashion, improving efficiency by overlapping computation and communication. Hybrid parallelism \cite{narayanan2021efficient} combines data, tensor, and pipeline parallelism to maximize resource utilization depending on the model's architecture and available hardware.

\section{Algorithm Details}
\label{sec: Algorithm Details}

We describe the full details of \sysname(\shortname) backward pass. Algorithm \ref{alg:stream_MLP_backward} shows the MLP backward, and Algorithm \ref{alg:stream_o_back} shows the LM-Head backward.

\begin{algorithm}
\caption{\label{alg:stream_MLP_backward}Mini-Sequence MLP Backward}
\begin{algorithmic}[1]
\REQUIRE Gradients of output $\nabla O \in \mathbb{R}^{N \times d}$, Matrices $X \in \mathbb{R}^{N \times d}$, Weights of three linear layers $W_{gate}, W_{up} \in \mathbb{R}^{d \times I}$, $W_{down} \in \mathbb{R}^{I \times d}$

\STATE Partition matrices $X$ into $M$ blocks $X_1, \dots, X_m$ of size $N_m \times d$, where $N_m = N/M$
\STATE Partition matrices $\nabla O$ into $M$ blocks $\nabla O_1, \ldots, \nabla O_m$ of size $N_m \times d$, where $N_m = N/M$

\FOR{$1 \leq i \leq M$}
\STATE Compute $\nabla X_i = \nabla MLP(\nabla O_i)$
\STATE Compute $\nabla W_{down}, \nabla W_{up}, \nabla W_{gate} += \nabla MLP_{gradient}(\nabla O_i, X_i)$
\ENDFOR
\STATE Concatenate $\nabla X = {\nabla X_1, \ldots, \nabla X_m} \in \mathbb{R}^{N \times d}$
\STATE Return $\nabla X$, $\nabla W_{gate}$, $\nabla W_{up}$, $\nabla W_{down}$.
\end{algorithmic}
\end{algorithm}

\begin{algorithm}
\caption{\label{alg:stream_o_back}Mini-Sequence LM-Head Backward}
\begin{algorithmic}[1]
\REQUIRE Loss gradients $\nabla loss \in \mathbb{R}^1$, Logits $\in \mathbb{R}^{N \times V}$, Labels $L \in \mathbb{R}^N$, Weights $W_{out} \in \mathbb{R}^{d \times V}$

\STATE Partition matrices $X$ into $M$ blocks $X_1, \dots, X_m$ of size $N_m \times d$, where $N_m = N/M$
\STATE Partition labels $L$ into $M$ sub-labels $L_1, \ldots, L_m$ of size $\frac{N}{m}$, where $\frac{N}{m} = \frac{N}{M}$
\STATE Activation Recomputation with backward
\FOR{$1 \leq i \leq M$}
\STATE Compute  $logits_{i} = X_i W_{out} $, $logits_{i} \in \mathbb{R}^{N_m \times V}$
\STATE Compute $\nabla logits_i = \text{CrossEntropyLossBackward}(Logits_i, L_i)$
\STATE Compute $\nabla X_i = \nabla logits_i W_{out}^T$, $\nabla X_i \in \mathbb{R}^{\frac{N}{m} \times d}$
\STATE Compute $\nabla W_{out} += X_i^T \nabla logits_i$
\STATE Compute $\nabla X_i = \nabla X_i \odot \nabla loss$
\STATE Compute $\nabla W_{out} = \nabla W_{out} \odot \nabla loss$
\ENDFOR 
\STATE Concatenate $\nabla X = {\nabla X_1, \ldots, \nabla X_m} \in \mathbb{R}^{N \times d}$
\STATE Return $\nabla X$, $\nabla W_{out}$.
\end{algorithmic}
\end{algorithm}

We now observe about \shortname backward pass that when computing the gradients of MLP and LM-Head, we do not need to use full input and intermediates data. Instead, we can use $1/M$ reduced data with mini-sequence, significantly reducing the intermediate value memory overhead.

The main idea of backward is to accumulate gradients $\nabla W$ generated from each mini-sequence $X_i$ as Algorithm \ref{alg:stream_MLP_backward} line 5 and Algorithm \ref{alg:stream_o_back} line 8. We get the exact same result as standard implementation by accumulating all mini-sequence gradients.

\section{Llama2 and Llama3: Model Architecture Comparison}
\label{sec: Model Architecture Comparison}

This appendix highlights the key architectural differences between the Llama2-7B and Llama3-8B models implemented by Hugging Face. The main distinction lies in the configuration of the MLP blocks and LM-Head (linear with cross-entropy loss) within the model architecture.

\begin{figure}[h]
\begin{center}
  \includegraphics[scale=0.45]{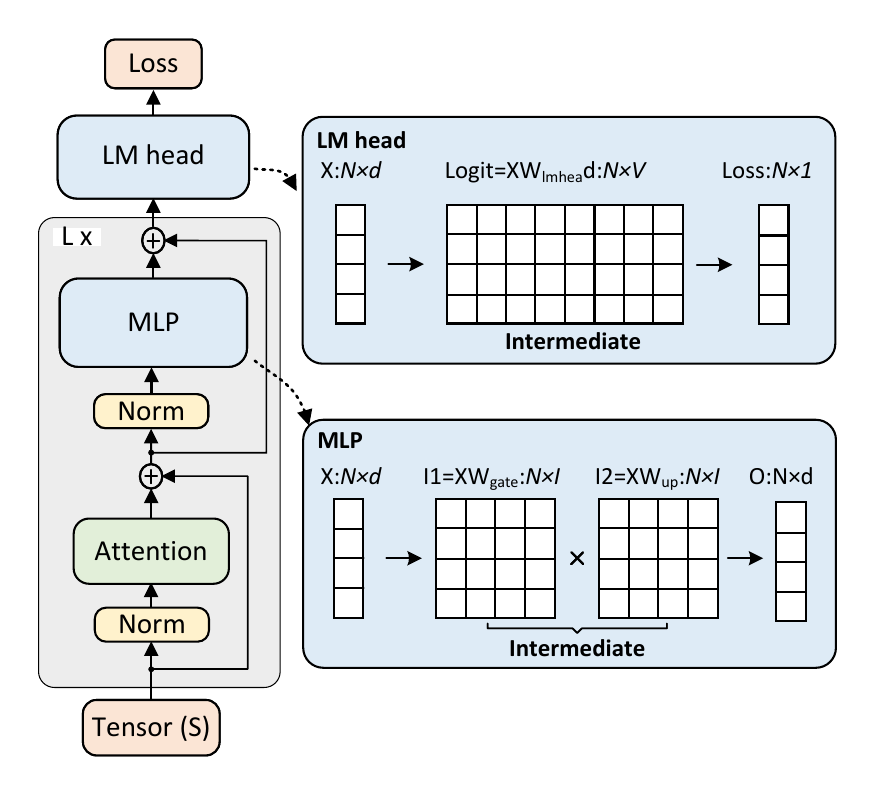}
  \vspace{-1em}
  \caption{Standard Transformer architecture with the highlight of LM-head and MLP.}
  \label{fig:transformer-highligh}
  \vspace{-1em}
\end{center}
\end{figure}

Figure \ref{fig:transformer-highligh} illustrates a standard Transformer model's architecture, focusing on the MLP and LM-head components. The intermediate tensors (I1, I2) have larger dimensions than the input and output. Also, the logit tensor has a much larger vocabulary dimension (V) than the hidden dimension (d).
\lstset{
    language=Python, 
    basicstyle=\ttfamily\tiny, 
    keywordstyle=\color{blue}, 
    commentstyle=\color{green}, 
    stringstyle=\color{red}, 
    numberstyle=\tiny\color{gray}, 
    stepnumber=1, 
    breaklines=true, 
    frame=single, 
    tabsize=4 
}

\captionof{figure}{Model Architecture of HuggingFace Implementation of Llama2-7B}
\label{fig:llama2_arch}
\lstset{language=Python, xleftmargin=1cm, xrightmargin=1cm}
\tiny
\begin{lstlisting}[frame=lines]{python}
LlamaForCausalLM(
  (model): LlamaModel(
    (embed_tokens): Embedding(32000, 4096)
    (layers): ModuleList(
      (0-31): 32 x LlamaDecoderLayer(
        (self_attn): LlamaFlashAttention2(
          (q_proj): Linear(in_features=4096, out_features=4096, bias=False)
          (k_proj): Linear(in_features=4096, out_features=4096, bias=False)
          (v_proj): Linear(in_features=4096, out_features=4096, bias=False)
          (o_proj): Linear(in_features=4096, out_features=4096, bias=False)
          (rotary_emb): LlamaRotaryEmbedding()
        )
        (mlp): LlamaMLP(
          (gate_proj): Linear(in_features=4096, out_features=11008, bias=False)
          (up_proj): Linear(in_features=4096, out_features=11008, bias=False)
          (down_proj): Linear(in_features=11008, out_features=4096, bias=False)
          (act_fn): SiLU()
        )
        (input_layernorm): LlamaRMSNorm((4096,), eps=1e-05)
        (post_attention_layernorm): LlamaRMSNorm((4096,), eps=1e-05)
      )
    )
    (norm): LlamaRMSNorm((4096,), eps=1e-05)
    (rotary_emb): LlamaRotaryEmbedding()
  )
  (lm_head): Linear(in_features=4096, out_features=32000, bias=False)
)
\end{lstlisting}
\normalsize



\paragraph{Llama2-7B Model Architecture:} As shown in Figure \ref{fig:llama2_arch}, the Llama2-7B model employs an MLP block configuration with the following characteristics:
\begin{itemize}
\item MLP Block:
The first linear layer projects the input from a hidden size of 4096 to an intermediate size of 11008.
The second linear layer projects the intermediate representation from 11008 to 4096, the hidden size.
\item LM-Head (Output Projection with Linear Loss):
The LM-Head in Llama2-7B consists of a linear layer that projects the hidden representation from a size of 4096 to a vocabulary size of 32000.
The output of the linear layer is then passed through a cross-entropy loss function to compute the training loss.

\end{itemize}

\captionof{figure}{Model Architecture of HuggingFace Implementation of Llama3-8B.}
\label{fig:llama3_arch}
\tiny
\lstset{language=Python, xleftmargin=1cm, xrightmargin=1cm}
\begin{lstlisting}[frame=lines]{python}
LlamaForCausalLM(
  (model): LlamaModel(
    (embed_tokens): Embedding(128256, 4096)
    (layers): ModuleList(
      (0-31): 32 x LlamaDecoderLayer(
        (self_attn): LlamaFlashAttention2(
          (q_proj): Linear(in_features=4096, out_features=4096, bias=False)
          (k_proj): Linear(in_features=4096, out_features=1024, bias=False)
          (v_proj): Linear(in_features=4096, out_features=1024, bias=False)
          (o_proj): Linear(in_features=4096, out_features=4096, bias=False)
          (rotary_emb): LlamaRotaryEmbedding()
        )
        (mlp): LlamaMLP(
          (gate_proj): Linear(in_features=4096, out_features=14336, bias=False)
          (up_proj): Linear(in_features=4096, out_features=14336, bias=False)
          (down_proj): Linear(in_features=14336, out_features=4096, bias=False)
          (act_fn): SiLU()
        )
        (input_layernorm): LlamaRMSNorm((4096,), eps=1e-05)
        (post_attention_layernorm): LlamaRMSNorm((4096,), eps=1e-05)
      )
    )
    (norm): LlamaRMSNorm((4096,), eps=1e-05)
    (rotary_emb): LlamaRotaryEmbedding()
  )
  (lm_head): Linear(in_features=4096, out_features=128256, bias=False)
)
\end{lstlisting}
\normalsize

\paragraph{Llama3-8B Model Architecture:} As shown in Figure \ref{fig:llama3_arch}, the Llama3-8B model employs an MLP block configuration with the following characteristics:

\begin{itemize}
\item MLP Block:
The first linear layer projects the input from a hidden size of 4096 to a larger intermediate size of 13824.
The second linear layer then projects the intermediate representation from a size of 13824 back to the hidden size of 4096.
\item LM-Head (Linear with Cross-Entropy Loss):
In Llama3-8B, the LM-Head also consists of a linear layer, but it projects the hidden representation from a size of 4096 to a larger vocabulary size of 32000.
Like Llama2-7B, the output of the linear layer is passed through a cross-entropy loss function to compute the training loss.
\end{itemize}

The increased intermediate size in the MLP block of Llama3-8B allows the model to capture complex patterns and transformations more effectively. Additionally, the larger vocabulary size in the LM-HEAD of Llama3-8B enables the model to generate more diverse and nuanced outputs.

It is worth noting that while the dimensions of the MLP block and LM-Head differ between Llama2-7B and Llama3-8B, the overall structure and functionality of these components remain the same. The cross-entropy loss function in the LM-Head measures the discrepancy between the predicted word probabilities and the target words during training, guiding the model to generate more accurate and contextually relevant outputs.
These architectural differences contribute to Llama3-8B's enhanced performance and capabilities compared to its predecessor, Llama2-7B, while maintaining a consistent overall structure. However, the unexpected large intermediate value also comes from the change of MLP blocks and LM-HEAD (linear with cross-entropy loss), which we will discuss in the following section.

Moreover, this trend is also reflected in the Microsoft development of Phi-3 \cite{abdin2024phi} compared with Phi-2 \cite{javaheripi2023phi}. Whose vocabulary size increased from 50k to 100K ($2\times$), intermediate size increased from 10k to 16k ($1.6\times$), and hidden size slightly increased from 2560 to 3072 ($1.2\times$). 

In conclusion, these models share an obvious trend: the ratio between intermediate and hidden size (also vocabulary and hidden size) is becoming larger.

\begin{figure}[h]
    \centering
  \includegraphics[width=1\linewidth]{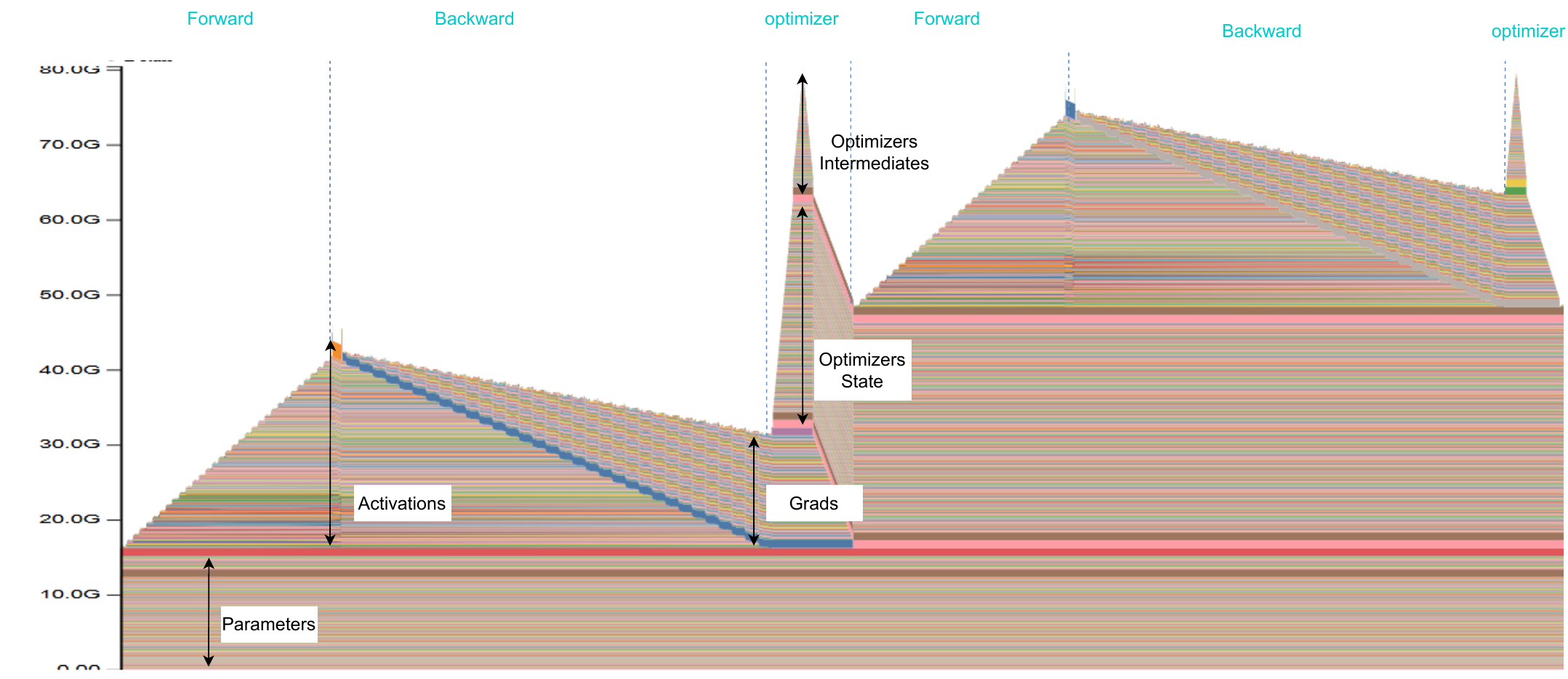}
  \caption{Memory Visualizaion of  training Llama3 8B with 4k sequence length on A100-80G.}
  \label{fig:mem_viz}
\vspace{-0.1in}
\end{figure}

\section{\sysname's  Memory Optimization Detials}
\label{sec:Memory Optimization Detials}
We compare \sysname(\shortname) with capturing and visualizing memory snapshots. We take vanilla Pytorch training for Llama3-8B with 4k sequence length as an example to show how the memory changes with the timeline.

\paragraph{vanilla.} Figure \ref{fig:mem_viz} shows the vanilla example. The model parameters had already been loaded into memory before the training step, so we immediately see a chunk of memory devoted to the weights. For Llama3-8B, its weight would be 15GB. As we start our forward pass, memory is allocated gradually for the activations or the tensors we are saving to be able to compute gradients in the backward pass. Here, the memory allocated for activation is larger than the weight, with around 29GB. Once we start the backward pass, the activations are gradually freed while the memory of the gradients starts building up. As the gradient is equal to the size of weights, which is smaller than activation, we can observe the memory usage drop to around 30 GB. Lastly, as the optimizer kicks in, its state will be lazily initialized, so we should see the optimizer state memory gradually increase during the optimizer step of the first training loop only. In future loops, the optimizer memory will remain and be updated. The memory for the gradients is then freed accordingly at the end of every training loop when it is called zero grade. Here, the optimizer would take $2\times$ of weights when using Adam with 30GB, and the optimizer intermediate is equal to the size of weight with 15. Therefore, the peak memory usage is during the optimizer step, which equals the sum size of weight, gradient, optimizer state, and optimizer intermediates, which roughly equals to $5\times$ of weight size with 75GB as shown in Table \ref{table:ap1}.

\begin{table}[h]
  \small
  \centering
  \caption{\label{table:ap1}Memory overhead of training Llama3-8B on single A100 80G GPU.
}
  {
    \begin{tabular}{@{}c|c|c@{}}
      Llama3-8B-hf & Memory Overhead & Within Peak Memory  \\ 
      \hline
      Activation  & 29 & 0\\
      Weight  & 15 & 15\\
      Gradient  & 15 & 15  \\
      Optimizer  & 45& 45  \\
      \hline
      Total  & - & 75  \\
      \hline
    \end{tabular}
  }
\end{table}

\begin{figure}[h]
    \centering
  \includegraphics[width=1\linewidth]{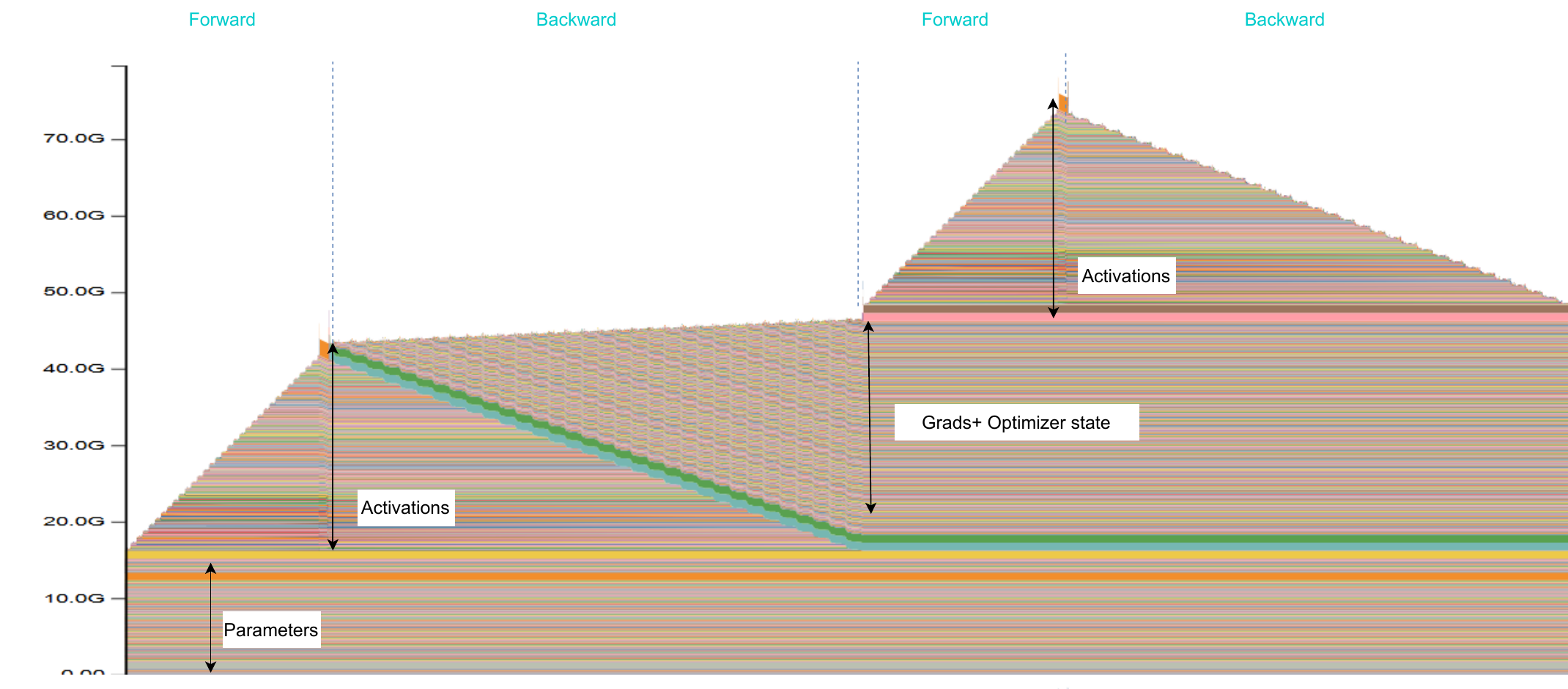}
  \caption{Memory Visualizaion of training Llama3 8B with 4k sequence length on A100-80G. The optimizer in the Backward technique is deployed here}
  \label{fig:mem_viz_optimizer_in_backward}
\vspace{-0.1in}
\end{figure}

\paragraph{optimizer-in-backward.} The first memory optimization discussed here is optimizer-in-backward \cite{optimizerstepinbackward}. It fuses the optimizer state with a backward pass to save the memory of the gradient and optimizer intermediate. The memory visualization of optimizer-in-backward is shown in Figure \ref{fig:mem_viz_optimizer_in_backward}, where there is no optimizer stage but only forward and backward state. The backward time would become larger as a result of fusion. Using this technology, the peak memory would change into the sum of weight, optimizer state, and activations. Although we successfully saved 30GB of memory overhead of gradient and optimizer intermediates, it adds up to 29GB memory overhead of activations, with only 1GB of memory saving. Totally it consumes 74GB of memory, as shown in Table \ref{table:ap2}. It would be worse if sequence length were increased, introducing more activation into LLM training. Therefore, optimizer-in-backward can hardly benefit long sequence training, but it simplifies the training process, so we would include this technique in the following discussion.

\begin{table}[h]
  \small
  \centering
  \caption{\label{table:ap2}Memory overhead of training Llama3-8B on single A100 80G GPU. The optimizer in the Backward technique is deployed.
}
  {
    \begin{tabular}{@{}c|c|c@{}}
      Llama3-8B-hf & Memory Overhead & Within Peak Memory  \\ 
      \hline
      Activation  & 29 & 29\\
      Weight  & 15 & 15\\
      Gradient  & 15 & 0  \\
      Optimizer  & 30 & 30  \\
      \hline
      Total  & - & 74  \\
      \hline
    \end{tabular}
  }
  \vspace{-1em}
\end{table}

\begin{figure}[h]
    \centering
  \includegraphics[width=1\linewidth]{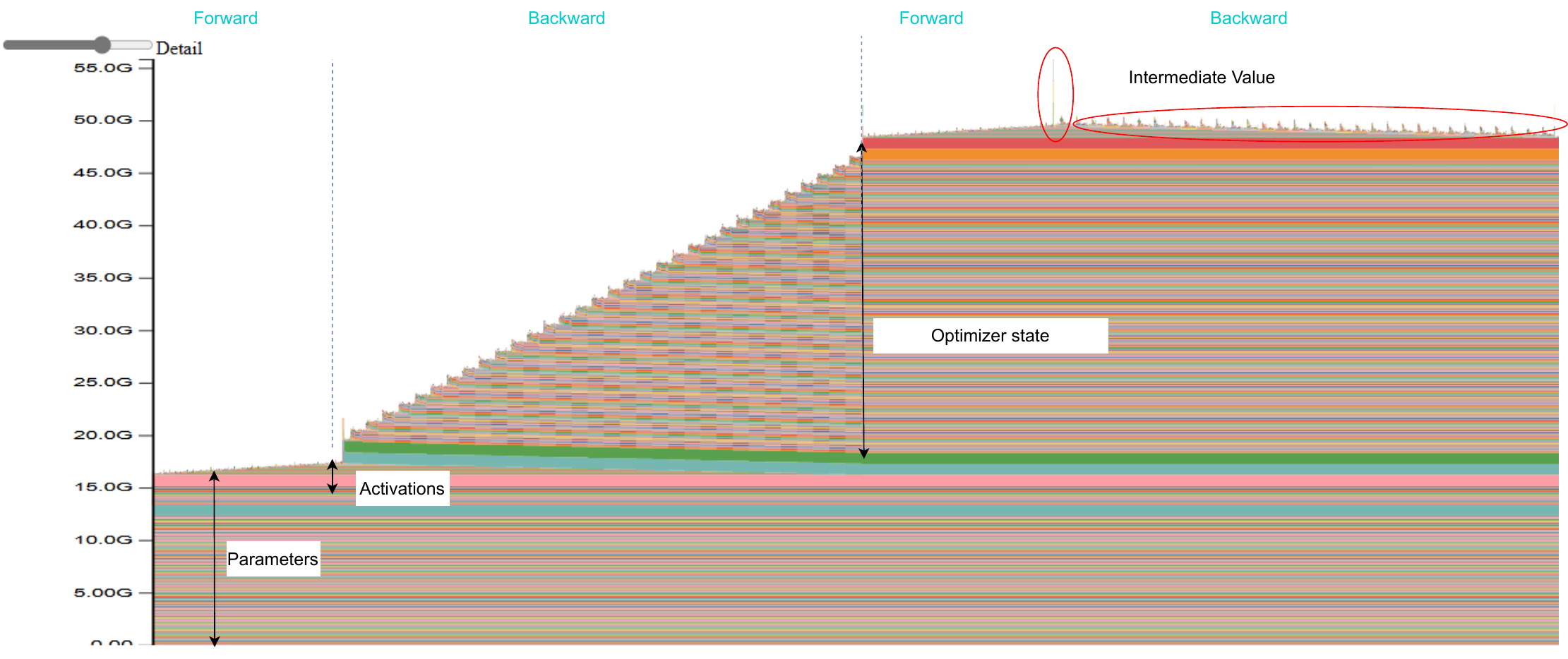}
  \caption{Memory Visualizaion of training Llama3 8B with 4k sequence length on A100-80G. Activation Recomputation technique is deployed here}
  \label{fig:mem_viz_activation_recomputation}
\vspace{-0.1in}
\end{figure}

\paragraph{Activation Recomputation.} Activation Recomputation is a powerful technique used in our analysis. It can significantly reduce the activation at the cost of a small decrease in the training speed due to recomputing parts of the graph during back-propagation. As shown in figure \ref{fig:mem_viz_activation_recomputation}, it successfully reduces the total memory overhead from 74GB to 52GB and reduces activation memory overhead from 29GB to 7GB with $4\times$ memory saving. However, we can easily find many singular points in the graph, which appear as an impulse signal. This impulse signal's duration is concise, meaning that it is intermediate data that is briefly created in the forward/backward pass and immediately removed from the GPU HBM memory. The most prominent intermediate data is several times the total activation (4-5 times in our data analysis). These intermediate data seriously affect the training performance of long sequences and become the activation bottleneck of the row.

\begin{table}[h]
  \small
  \centering
  \caption{\label{table:ap3}Memory overhead of training Llama3-8B on single A100 80G GPU. Activation Recomputation technique is deployed.
}
  {
    \begin{tabular}{@{}c|c|c@{}}
      Llama3-8B-hf & Memory Overhead & Within Peak Memory  \\ 
      \hline
      Activation  & 7 & 7\\
      Weight  & 15 & 15\\
      Gradient  & 15 & 0  \\
      Optimizer  & 30 & 30  \\
      \hline
      Total  & - & 52  \\
      \hline
    \end{tabular}
  }
  \vspace{-1em}
\end{table}

\begin{figure}[h]
    \centering
  \includegraphics[width=1\linewidth]{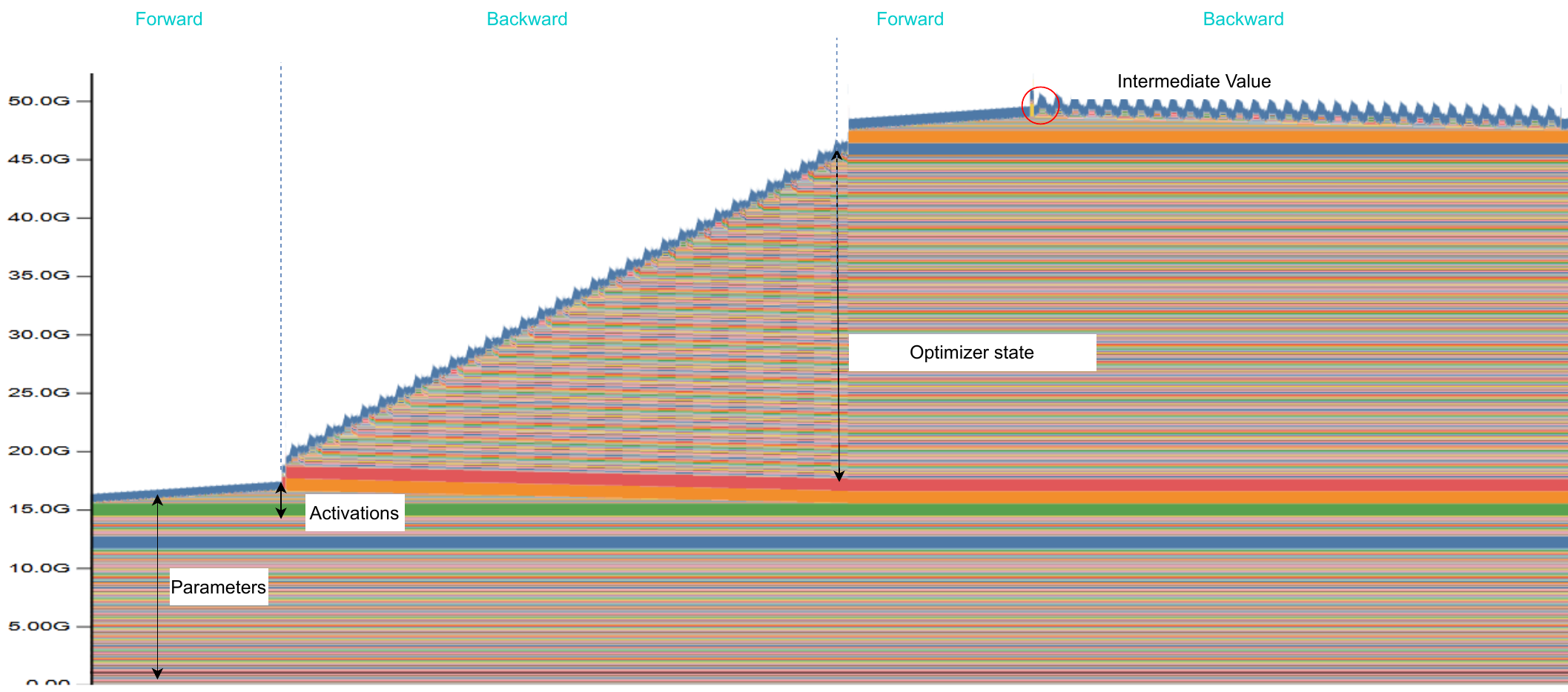}
  \caption{Memory Visualizaion of training Llama3 8B with 4k sequence length on A100-80G. \sysname technique is deployed here}
  \label{fig:mem_viz_narrow}
\vspace{-0.1in}
\end{figure}

\paragraph{\sysname(\shortname)} Inspired by the observation from Activation Recomputation techniques, we propose \shortname to reduce the intermediate value during training. We successfully decrease the memory overhead of activation from 7GB to 4GB, while the intermediate value is significantly reduced. This is because only $1/M$ intermediate values are used for computing the forward outputs, backward errors, and gradients during both the forward and backward.

\begin{table}[h]
  \small
  \centering
  \caption{\label{table:ap4}Memory overhead of training Llama3-8B on single A100 80G GPU. \sysname technique is deployed.
}
  {
    \begin{tabular}{@{}c|c|c@{}}
      Llama3-8B-hf & Memory Overhead & Within Peak Memory  \\ 
      \hline
      Activation  & 2 & 2\\
      Weight  & 15 & 15\\
      Gradient  & 15 & 0  \\
      Optimizer  & 30 & 30  \\
      \hline
      Total  & - & 48  \\
      \hline
    \end{tabular}
  }
  \vspace{-1em}
\end{table}

\begin{figure}[h]
    \centering
  \includegraphics[width=0.8\linewidth]{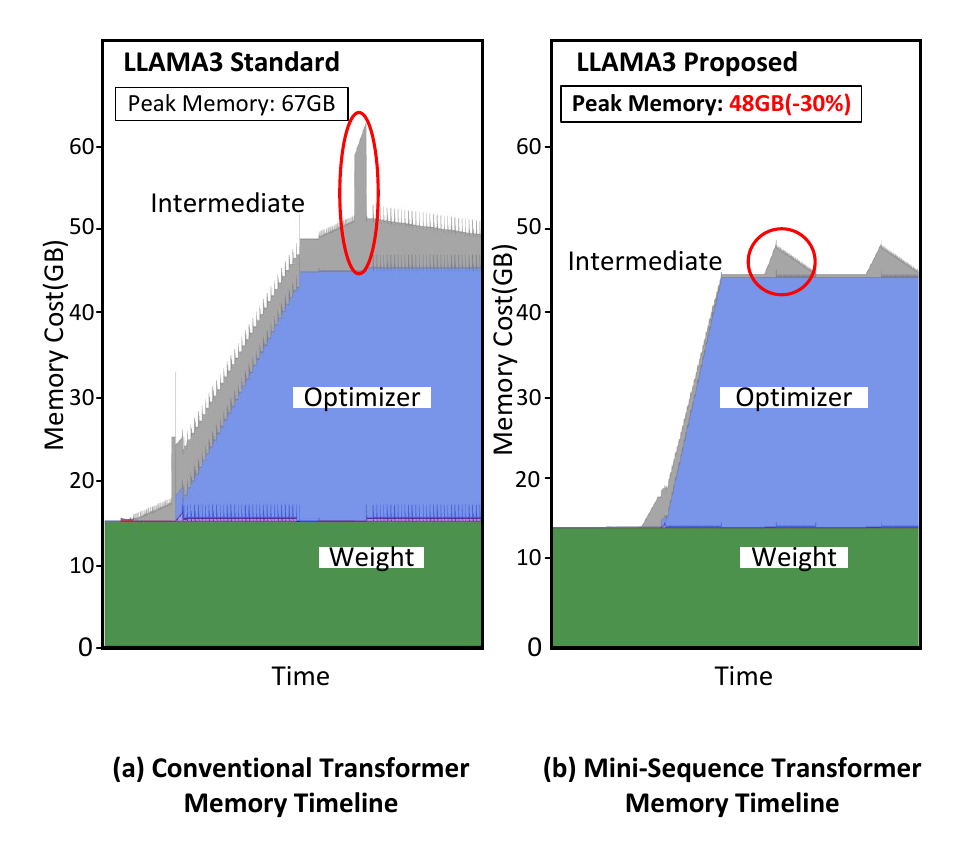}
  \caption{Memory Visualization. (a) The memory timeline of training Llama3-8B using standard transformer architecture, Red cycle highlights the intermediate memory (b) The memory timeline of training Llama3-8B using \shortname, Red cycle highlights the intermediate memory has been narrowed.}
  \label{fig:com}
\vspace{-0.1in}
\end{figure}

\paragraph{Conclusion.} we compare the memory usage over time when training the Llama3-8B model using the standard transformer architecture versus using \shortname, which is shown on \ref{fig:com}.

In Figure \ref{fig:com}(a), which shows the memory timeline for the standard Llama3-8B training, we can see that the memory usage is dominated by three main components: the model weights (in blue), the optimizer state (in green), and the intermediate memory (highlighted by the red circle). The peak memory usage reaches around 67GB.

In contrast, Figure \ref{fig:com}(b) demonstrates the memory timeline when training Llama3-8B with \shortname. The critical difference is that the intermediate memory, again highlighted by the red circle, has been significantly reduced or "narrowed" compared to the standard transformer case. As a result, the peak memory usage with \shortname is around 47GB, achieving a $30\%$ reduction compared to the standard transformer.

The memory timelines illustrate that \shortname effectively reduces the intermediate memory footprint during training, significantly contributing to overall memory consumption. By minimizing the intermediate memory, \shortname enables more efficient memory utilization and allows training with longer sequence lengths or larger batch sizes while staying within the hardware's available memory limits.


\section{Scaling to Extreme Long Sequence on Distributed Setting}
\label{distributed experiment}

We evaluate our distributed extension of \shortname on Llama3-8B Llama2-7B models and compare against vanilla DeepSpeed-Ulysses‘s sequence parallelism on 2,4,8 GPUs, respectively, for the max sequence length and corresponding training time. The results of this evaluation are shown in Table \ref{table:distributed}. 

\begin{table}[h]
  \small
  \centering
  \caption{\label{table:distributed}Maximum sequence length of Llama3-8B, running on distributed setting.
}
  {
    \begin{tabular}{@{}c|c|c|c@{}}
      Model Implementation & \multicolumn{3}{|c}{GPU numbers}  \\ 
       & 2 & 4 &8  \\ \hline
      Llama3-8b-hf \shortname   & 120 & 240 & 480    \\
      Llama2-7B-hf \shortname  & 160 & 320 & 640  \\
       \hline
    \end{tabular}
  }
  \vspace{-1em}
\end{table}

\section{How many mini-sequences are needed during pre-training}
\label{M number}

 We also provide insights into the performance characteristics and memory usage of the Llama3-8B model when using the \sysname(\shortname) approach with different numbers of mini-sequences (M) and sequence lengths.

Table \ref{tab:lm_head_time} shows the execution time for different sequence lengths and mini-sequence settings. As the number of mini-sequences (M) increases, the execution time slightly increases, especially for shorter sequences. However, for longer sequences (e.g., 80000), the execution time remains relatively stable across different mini-sequences (M).

\begin{table}[h]
\centering

\caption{LM-head time for different sequence lengths and mini-sequence settings}
\begin{tabular}{c|c|c|c|c|c|c|c}
\hline
LM-head time & 1024 & 2048 & 4096 & 8192 & 20000 & 40000 & 80000 \\
\hline
standard & 0.01 & 0.02 & 0.04 & 0.09 & 0.2 & 0.4 & 0.85 \\
\hline
M=2 & 0.02 & 0.04 & 0.07 & 0.14 & 0.31 & 0.67 & 1.36 \\

M=4 & 0.03 & 0.04 & 0.07 & 0.14 & 0.33 & 0.67 & 1.33 \\

M=8 & 0.04 & 0.05 & 0.08 & 0.14 & 0.34 & 0.67 & 1.34 \\

M=16 & 0.06 & 0.07 & 0.09 & 0.16 & 0.34 & 0.68 & 1.35 \\

M=32 & 0.11 & 0.11 & 0.14 & 0.19 & 0.37 & 0.69 & 1.36 \\
\hline
\end{tabular}
\label{tab:lm_head_time}
\end{table}

Table \ref{tab:lm_head_memory} shows the memory usage in gigabytes (GB) for different sequence lengths and mini-sequence settings for the LM-Head component. The standard setting without mini-sequences consumes 59.92 GB for a sequence length of 80000. By increasing the number of mini-sequences, memory usage decreases significantly. With M=16, the memory usage reduces to 9.12 GB for the same sequence length, achieving an 84.8\% reduction in memory consumption. This is further improved with M=32 to 89.8\%.

\begin{table}[h]
\centering
\caption{LM-head memory usage (in GB) for different sequence lengths and mini-sequence settings}
\begin{tabular}{c|c|c|c|c|c|c|c}
\hline
LM-head memory & 1024 & 2048 & 4096 & 8192 & 20000 & 40000 & 80000 \\
\hline
standard & 3.20 & 3.46 & 4.94 & 7.91 & 16.46 & 30.95 & 59.92 \\
\hline
mini-seq=2 & 2.59 & 3.21 & 4.46 & 6.95 & 14.14 & 26.31 & 50.66 \\
mini-seq=4 & 2.28 & 2.60 & 3.24 & 4.52 & 8.20 & 14.44 & 26.92 \\
mini-seq=8 & 2.13 & 2.30 & 2.63 & 3.30 & 5.24 & 8.51 & 15.05 \\
mini-seq=16 & 2.06 & 2.15 & 2.33 & 2.70 & 4.14 & 5.54 & 9.12 \\
mini-seq=32 & 2.02 & 2.07 & 2.18 & 2.39 & 3.01 & 4.06 & 6.15 \\
\hline
\end{tabular}
\label{tab:lm_head_memory}
\end{table}

Table \ref{tab:mlp_time} shows the execution time for different sequence lengths and mini-sequence settings. Like the LM-Head, increasing M leads to slightly longer execution times, particularly for shorter sequences. However, the impact on execution time is minimal for longer sequences (e.g., 80000).

\begin{table}[h]
\centering
\caption{MLP time (in seconds) for different sequence lengths and mini-sequence settings}
\begin{tabular}{c|c|c|c|c|c|c|c}
\hline
MLP time & 1024 & 2048 & 4096 & 8192 & 20000 & 40000 & 80000 \\
\hline
standard & 0.05 & 0.08 & 0.16 & 0.31 & 0.74 & 1.52 & 2.96 \\
\hline
M=2 & 0.05 & 0.10 & 0.17 & 0.32 & 0.76 & 1.49 & 3.05 \\
M=4 & 0.07 & 0.11 & 0.19 & 0.33 & 0.79 & 1.52 & 2.99 \\
M=8 & 0.12 & 0.15 & 0.22 & 0.38 & 0.81 & 1.58 & 3.05 \\
\hline
\end{tabular}
\label{tab:mlp_time}
\end{table}

For the MLP component, Table \ref{tab:mlp_memory} demonstrates the memory usage for different sequence lengths and mini-sequence settings. The standard setting consumes 14.72 GB for a sequence length of 80000, while using M=8 mini-sequences reduces the memory usage to 11.66 GB, resulting in a 20.8\% reduction. Here, we can observe.

\begin{table}[h]
\centering
\caption{MLP memory usage (in GB) for different sequence lengths and mini-sequence settings}
\begin{tabular}{c|c|c|c|c|c|c|c}
\hline
MLP memory & 1024 & 2048 & 4096 & 8192 & 20000 & 40000 & 80000 \\
\hline
standard & 0.93 & 1.09 & 1.39 & 2.11 & 4.18 & 7.69 & 14.72 \\
\hline
M=2 & 1.29 & 1.36 & 1.50 & 2.00 & 3.76 & 6.73 & 12.69 \\
M=4 & 1.32 & 1.41 & 1.61 & 2.00 & 3.49 & 6.21 & 11.66 \\
M=8 & 1.33 & 1.44 & 1.66 & 2.11 & 3.42 & 6.17 & 11.66 \\
\hline
\end{tabular}
\label{tab:mlp_memory}
\end{table}

The analysis suggests that increasing the number of mini-sequences (M) can significantly reduce memory usage, especially for the LM-Head component, while having a minimal impact on execution time for longer sequences. Memory savings are more pronounced for the LM-Head than for the MLP.

It is important to note that while \shortname is highly beneficial for training with extremely long sequences, it may lead to performance degradation when applied to models with shorter sequences due to the overhead introduced by partitioning the input and the additional memory movement required for gradient accumulation. It can be easily observed from Table \ref{tab:lm_head_time} that using M=32 mini-sequences increases the execution time from 0.01s (standard setting) to 0.11s for a sequence length of 1024, causing an 11x performance downgrade. Also, from Table \ref{tab:mlp_time}, using M=8 mini-sequences increases the execution time from 0.05s (standard setting) to 0.12s for a sequence length of 1024, causing a 2x performance downgrade. The performance reduction is more pronounced for the LM-Head compared to the MLP. Fortunately, LM-head accounts for very little of the transformer's running time, which is smaller than MLP and much smaller than attention, so our technology will not affect the overall performance, even if it affects its module performance.

\section{Integrated with existing frameworks }
\label{Intergration}

MST's core idea is conceptually straightforward, primarily targeting MLP and LM-Head blocks. We offer two integration methods:




\paragraph{Customized Hugging Face Transformer.} This method involves directly modifying the Hugging Face Transformer library to incorporate MST functionality. By customizing the library, users can seamlessly integrate MST into their existing workflows that use Hugging Face Transformers. We made this method open-source on \href{https://github.com/wdlctc/transformers}{https://github.com/wdlctc/transformers}.

To use the customized Hugging Face Transformer with MST, ML developer didn't change any line but install our customized transformers library with \shortname:
\begin{verbatim}
import transformers
\end{verbatim}

\paragraph{Wrapper Mode.} The Wrapper Mode provides a less invasive approach to integrating MST. This method involves creating a wrapper around existing model implementations, intercepting and modifying the forward and backward passes of the MLP and LM-Head blocks. We made this method open-source on \href{https://github.com/wdlctc/mini-s}{https://github.com/wdlctc/mini-s}.

To use the Wrapper Mode:
\begin{verbatim}
from mini-s import mst
model = mst(model)
\end{verbatim}

\paragraph{Conclusion.} Both integration methods offer flexibility in adopting MST for long sequence training. The choice between Customized Hugging Face Transformer and Wrapper Mode depends on the specific requirements of the project, the level of integration desired, and the willingness to maintain custom libraries.

For users deeply invested in the Hugging Face ecosystem, the Customized Hugging Face Transformer method may be preferable. This method requires minimal changes to existing codebases which is already integrated with the  Hugging Face ecosystem, and allows access to all Hugging Face features and optimizations. For those seeking a more flexible solution or working with multiple model implementations, the Wrapper Mode could be the better choice to used with customized codebase challenges.

\newpage
\newpage
\section*{NeurIPS Paper Checklist}

The checklist is designed to encourage best practices for responsible machine learning research, addressing issues of reproducibility, transparency, research ethics, and societal impact. Do not remove the checklist: {\bf The papers not including the checklist will be desk rejected.} The checklist should follow the references and follow the (optional) supplemental material.  The checklist does NOT count towards the page
limit. 

Please read the checklist guidelines carefully for information on how to answer these questions. For each question in the checklist:
\begin{itemize}
    \item You should answer \answerYes{}, \answerNo{}, or \answerNA{}.
    \item \answerNA{} means either that the question is Not Applicable for that particular paper or the relevant information is Not Available.
    \item Please provide a short (1–2 sentence) justification right after your answer (even for NA). 
\end{itemize}

{\bf The checklist answers are an integral part of your paper submission.} They are visible to the reviewers, area chairs, senior area chairs, and ethics reviewers. You will be asked to also include it (after eventual revisions) with the final version of your paper, and its final version will be published with the paper.

The reviewers of your paper will be asked to use the checklist as one of the factors in their evaluation. While "\answerYes{}" is generally preferable to "\answerNo{}", it is perfectly acceptable to answer "\answerNo{}" provided a proper justification is given (e.g., "error bars are not reported because it would be too computationally expensive" or "we were unable to find the license for the dataset we used"). In general, answering "\answerNo{}" or "\answerNA{}" is not grounds for rejection. While the questions are phrased in a binary way, we acknowledge that the true answer is often more nuanced, so please just use your best judgment and write a justification to elaborate. All supporting evidence can appear either in the main paper or the supplemental material, provided in appendix. If you answer \answerYes{} to a question, in the justification please point to the section(s) where related material for the question can be found.

IMPORTANT, please:
\begin{itemize}
    \item {\bf Delete this instruction block, but keep the section heading ``NeurIPS paper checklist"},
    \item  {\bf Keep the checklist subsection headings, questions/answers and guidelines below.}
    \item {\bf Do not modify the questions and only use the provided macros for your answers}.
\end{itemize}


\begin{enumerate}

\item {\bf Claims}
    \item[] Question: Do the main claims made in the abstract and introduction accurately reflect the paper's contributions and scope?
    \item[] Answer: \answerYes{} 
    \item[] Justification: Section \ref{Introduction}
    \item[] Guidelines:
    \begin{itemize}
        \item The answer NA means that the abstract and introduction do not include the claims made in the paper.
        \item The abstract and/or introduction should clearly state the claims made, including the contributions made in the paper and important assumptions and limitations. A No or NA answer to this question will not be perceived well by the reviewers. 
        \item The claims made should match theoretical and experimental results, and reflect how much the results can be expected to generalize to other settings. 
        \item It is fine to include aspirational goals as motivation as long as it is clear that these goals are not attained by the paper. 
    \end{itemize}

\item {\bf Limitations}
    \item[] Question: Does the paper discuss the limitations of the work performed by the authors?
    \item[] Answer: \answerYes{} 
    \item[] Justification: Section \ref{sec:discussion}
    \item[] Guidelines:
    \begin{itemize}
        \item The answer NA means that the paper has no limitation while the answer No means that the paper has limitations, but those are not discussed in the paper. 
        \item The authors are encouraged to create a separate "Limitations" section in their paper.
        \item The paper should point out any strong assumptions and how robust the results are to violations of these assumptions (e.g., independence assumptions, noiseless settings, model well-specification, asymptotic approximations only holding locally). The authors should reflect on how these assumptions might be violated in practice and what the implications would be.
        \item The authors should reflect on the scope of the claims made, e.g., if the approach was only tested on a few datasets or with a few runs. In general, empirical results often depend on implicit assumptions, which should be articulated.
        \item The authors should reflect on the factors that influence the performance of the approach. For example, a facial recognition algorithm may perform poorly when image resolution is low or images are taken in low lighting. Or a speech-to-text system might not be used reliably to provide closed captions for online lectures because it fails to handle technical jargon.
        \item The authors should discuss the computational efficiency of the proposed algorithms and how they scale with dataset size.
        \item If applicable, the authors should discuss possible limitations of their approach to address problems of privacy and fairness.
        \item While the authors might fear that complete honesty about limitations might be used by reviewers as grounds for rejection, a worse outcome might be that reviewers discover limitations that aren't acknowledged in the paper. The authors should use their best judgment and recognize that individual actions in favor of transparency play an important role in developing norms that preserve the integrity of the community. Reviewers will be specifically instructed to not penalize honesty concerning limitations.
    \end{itemize}

\item {\bf Theory Assumptions and Proofs}
    \item[] Question: For each theoretical result, does the paper provide the full set of assumptions and a complete (and correct) proof?
    \item[] Answer: \answerYes{} 
    \item[] Justification: Section \ref{Analysis}
    \item[] Guidelines:
    \begin{itemize}
        \item The answer NA means that the paper does not include theoretical results. 
        \item All the theorems, formulas, and proofs in the paper should be numbered and cross-referenced.
        \item All assumptions should be clearly stated or referenced in the statement of any theorems.
        \item The proofs can either appear in the main paper or the supplemental material, but if they appear in the supplemental material, the authors are encouraged to provide a short proof sketch to provide intuition. 
        \item Inversely, any informal proof provided in the core of the paper should be complemented by formal proofs provided in appendix or supplemental material.
        \item Theorems and Lemmas that the proof relies upon should be properly referenced. 
    \end{itemize}

    \item {\bf Experimental Result Reproducibility}
    \item[] Question: Does the paper fully disclose all the information needed to reproduce the main experimental results of the paper to the extent that it affects the main claims and/or conclusions of the paper (regardless of whether the code and data are provided or not)?
    \item[] Answer: \answerYes{} 
    \item[] Justification: Appendix \ref{sec:Memory Optimization Detials}
    \item[] Guidelines:
    \begin{itemize}
        \item The answer NA means that the paper does not include experiments.
        \item If the paper includes experiments, a No answer to this question will not be perceived well by the reviewers: Making the paper reproducible is important, regardless of whether the code and data are provided or not.
        \item If the contribution is a dataset and/or model, the authors should describe the steps taken to make their results reproducible or verifiable. 
        \item Depending on the contribution, reproducibility can be accomplished in various ways. For example, if the contribution is a novel architecture, describing the architecture fully might suffice, or if the contribution is a specific model and empirical evaluation, it may be necessary to either make it possible for others to replicate the model with the same dataset, or provide access to the model. In general. releasing code and data is often one good way to accomplish this, but reproducibility can also be provided via detailed instructions for how to replicate the results, access to a hosted model (e.g., in the case of a large language model), releasing of a model checkpoint, or other means that are appropriate to the research performed.
        \item While NeurIPS does not require releasing code, the conference does require all submissions to provide some reasonable avenue for reproducibility, which may depend on the nature of the contribution. For example
        \begin{enumerate}
            \item If the contribution is primarily a new algorithm, the paper should make it clear how to reproduce that algorithm.
            \item If the contribution is primarily a new model architecture, the paper should describe the architecture clearly and fully.
            \item If the contribution is a new model (e.g., a large language model), then there should either be a way to access this model for reproducing the results or a way to reproduce the model (e.g., with an open-source dataset or instructions for how to construct the dataset).
            \item We recognize that reproducibility may be tricky in some cases, in which case authors are welcome to describe the particular way they provide for reproducibility. In the case of closed-source models, it may be that access to the model is limited in some way (e.g., to registered users), but it should be possible for other researchers to have some path to reproducing or verifying the results.
        \end{enumerate}
    \end{itemize}

\item {\bf Open access to data and code}
    \item[] Question: Does the paper provide open access to the data and code, with sufficient instructions to faithfully reproduce the main experimental results, as described in supplemental material?
    \item[] Answer: \answerYes{} 
    \item[] Justification: supply meterials
    \item[] Guidelines:
    \begin{itemize}
        \item The answer NA means that paper does not include experiments requiring code.
        \item Please see the NeurIPS code and data submission guidelines (\url{https://nips.cc/public/guides/CodeSubmissionPolicy}) for more details.
        \item While we encourage the release of code and data, we understand that this might not be possible, so “No” is an acceptable answer. Papers cannot be rejected simply for not including code, unless this is central to the contribution (e.g., for a new open-source benchmark).
        \item The instructions should contain the exact command and environment needed to run to reproduce the results. See the NeurIPS code and data submission guidelines (\url{https://nips.cc/public/guides/CodeSubmissionPolicy}) for more details.
        \item The authors should provide instructions on data access and preparation, including how to access the raw data, preprocessed data, intermediate data, and generated data, etc.
        \item The authors should provide scripts to reproduce all experimental results for the new proposed method and baselines. If only a subset of experiments are reproducible, they should state which ones are omitted from the script and why.
        \item At submission time, to preserve anonymity, the authors should release anonymized versions (if applicable).
        \item Providing as much information as possible in supplemental material (appended to the paper) is recommended, but including URLs to data and code is permitted.
    \end{itemize}

\item {\bf Experimental Setting/Details}
    \item[] Question: Does the paper specify all the training and test details (e.g., data splits, hyperparameters, how they were chosen, type of optimizer, etc.) necessary to understand the results?
    \item[] Answer: \answerYes{} 
    \item[] Justification: Section \ref{others}
    \item[] Guidelines:
    \begin{itemize}
        \item The answer NA means that the paper does not include experiments.
        \item The experimental setting should be presented in the core of the paper to a level of detail that is necessary to appreciate the results and make sense of them.
        \item The full details can be provided either with the code, in appendix, or as supplemental material.
    \end{itemize}

\item {\bf Experiment Statistical Significance}
    \item[] Question: Does the paper report error bars suitably and correctly defined or other appropriate information about the statistical significance of the experiments?
    \item[] Answer: \answerNo{} 
    \item[] Justification:  Error bars are not reported because it would be too computationally exxpensive.
    \item[] Guidelines:
    \begin{itemize}
        \item The answer NA means that the paper does not include experiments.
        \item The authors should answer "Yes" if the results are accompanied by error bars, confidence intervals, or statistical significance tests, at least for the experiments that support the main claims of the paper.
        \item The factors of variability that the error bars are capturing should be clearly stated (for example, train/test split, initialization, random drawing of some parameter, or overall run with given experimental conditions).
        \item The method for calculating the error bars should be explained (closed form formula, call to a library function, bootstrap, etc.)
        \item The assumptions made should be given (e.g., Normally distributed errors).
        \item It should be clear whether the error bar is the standard deviation or the standard error of the mean.
        \item It is OK to report 1-sigma error bars, but one should state it. The authors should preferably report a 2-sigma error bar than state that they have a 96\% CI, if the hypothesis of Normality of errors is not verified.
        \item For asymmetric distributions, the authors should be careful not to show in tables or figures symmetric error bars that would yield results that are out of range (e.g. negative error rates).
        \item If error bars are reported in tables or plots, The authors should explain in the text how they were calculated and reference the corresponding figures or tables in the text.
    \end{itemize}

\item {\bf Experiments Compute Resources}
    \item[] Question: For each experiment, does the paper provide sufficient information on the computer resources (type of compute workers, memory, time of execution) needed to reproduce the experiments?
    \item[] Answer:\answerYes{} 
    \item[] Justification: A100 GPUs
    \item[] Guidelines:
    \begin{itemize}
        \item The answer NA means that the paper does not include experiments.
        \item The paper should indicate the type of compute workers CPU or GPU, internal cluster, or cloud provider, including relevant memory and storage.
        \item The paper should provide the amount of compute required for each of the individual experimental runs as well as estimate the total compute. 
        \item The paper should disclose whether the full research project required more compute than the experiments reported in the paper (e.g., preliminary or failed experiments that didn't make it into the paper). 
    \end{itemize}
    
\item {\bf Code Of Ethics}
    \item[] Question: Does the research conducted in the paper conform, in every respect, with the NeurIPS Code of Ethics \url{https://neurips.cc/public/EthicsGuidelines}?
    \item[] Answer: \answerYes{} 
    \item[] Justification: 
    \item[] Guidelines:
    \begin{itemize}
        \item The answer NA means that the authors have not reviewed the NeurIPS Code of Ethics.
        \item If the authors answer No, they should explain the special circumstances that require a deviation from the Code of Ethics.
        \item The authors should make sure to preserve anonymity (e.g., if there is a special consideration due to laws or regulations in their jurisdiction).
    \end{itemize}

\item {\bf Broader Impacts}
    \item[] Question: Does the paper discuss both potential positive societal impacts and negative societal impacts of the work performed?
    \item[] Answer: \answerNA{} 
    \item[] Justification: 
    \item[] Guidelines:
    \begin{itemize}
        \item The answer NA means that there is no societal impact of the work performed.
        \item If the authors answer NA or No, they should explain why their work has no societal impact or why the paper does not address societal impact.
        \item Examples of negative societal impacts include potential malicious or unintended uses (e.g., disinformation, generating fake profiles, surveillance), fairness considerations (e.g., deployment of technologies that could make decisions that unfairly impact specific groups), privacy considerations, and security considerations.
        \item The conference expects that many papers will be foundational research and not tied to particular applications, let alone deployments. However, if there is a direct path to any negative applications, the authors should point it out. For example, it is legitimate to point out that an improvement in the quality of generative models could be used to generate deepfakes for disinformation. On the other hand, it is not needed to point out that a generic algorithm for optimizing neural networks could enable people to train models that generate Deepfakes faster.
        \item The authors should consider possible harms that could arise when the technology is being used as intended and functioning correctly, harms that could arise when the technology is being used as intended but gives incorrect results, and harms following from (intentional or unintentional) misuse of the technology.
        \item If there are negative societal impacts, the authors could also discuss possible mitigation strategies (e.g., gated release of models, providing defenses in addition to attacks, mechanisms for monitoring misuse, mechanisms to monitor how a system learns from feedback over time, improving the efficiency and accessibility of ML).
    \end{itemize}
    
\item {\bf Safeguards}
    \item[] Question: Does the paper describe safeguards that have been put in place for responsible release of data or models that have a high risk for misuse (e.g., pretrained language models, image generators, or scraped datasets)?
    \item[] Answer: \answerNA{} 
    \item[] Justification: 
    \item[] Guidelines:
    \begin{itemize}
        \item The answer NA means that the paper poses no such risks.
        \item Released models that have a high risk for misuse or dual-use should be released with necessary safeguards to allow for controlled use of the model, for example by requiring that users adhere to usage guidelines or restrictions to access the model or implementing safety filters. 
        \item Datasets that have been scraped from the Internet could pose safety risks. The authors should describe how they avoided releasing unsafe images.
        \item We recognize that providing effective safeguards is challenging, and many papers do not require this, but we encourage authors to take this into account and make a best faith effort.
    \end{itemize}

\item {\bf Licenses for existing assets}
    \item[] Question: Are the creators or original owners of assets (e.g., code, data, models), used in the paper, properly credited and are the license and terms of use explicitly mentioned and properly respected?
    \item[] Answer: \answerYes{} 
    \item[] Justification: References
    \item[] Guidelines: 
    \begin{itemize}
        \item The answer NA means that the paper does not use existing assets.
        \item The authors should cite the original paper that produced the code package or dataset.
        \item The authors should state which version of the asset is used and, if possible, include a URL.
        \item The name of the license (e.g., CC-BY 4.0) should be included for each asset.
        \item For scraped data from a particular source (e.g., website), the copyright and terms of service of that source should be provided.
        \item If assets are released, the license, copyright information, and terms of use in the package should be provided. For popular datasets, \url{paperswithcode.com/datasets} has curated licenses for some datasets. Their licensing guide can help determine the license of a dataset.
        \item For existing datasets that are re-packaged, both the original license and the license of the derived asset (if it has changed) should be provided.
        \item If this information is not available online, the authors are encouraged to reach out to the asset's creators.
    \end{itemize}

\item {\bf New Assets}
    \item[] Question: Are new assets introduced in the paper well documented and is the documentation provided alongside the assets?
    \item[] Answer: \answerYes{} 
    \item[] Justification: \answerNA{}.
    \item[] Guidelines:
    \begin{itemize}
        \item The answer NA means that the paper does not release new assets.
        \item Researchers should communicate the details of the dataset/code/model as part of their submissions via structured templates. This includes details about training, license, limitations, etc. 
        \item The paper should discuss whether and how consent was obtained from people whose asset is used.
        \item At submission time, remember to anonymize your assets (if applicable). You can either create an anonymized URL or include an anonymized zip file.
    \end{itemize}

\item {\bf Crowdsourcing and Research with Human Subjects}
    \item[] Question: For crowdsourcing experiments and research with human subjects, does the paper include the full text of instructions given to participants and screenshots, if applicable, as well as details about compensation (if any)? 
    \item[] Answer: \answerNA{} 
    \item[] Justification: 
    \item[] Guidelines:
    \begin{itemize}
        \item The answer NA means that the paper does not involve crowdsourcing nor research with human subjects.
        \item Including this information in the supplemental material is fine, but if the main contribution of the paper involves human subjects, then as much detail as possible should be included in the main paper. 
        \item According to the NeurIPS Code of Ethics, workers involved in data collection, curation, or other labor should be paid at least the minimum wage in the country of the data collector. 
    \end{itemize}

\item {\bf Institutional Review Board (IRB) Approvals or Equivalent for Research with Human Subjects}
    \item[] Question: Does the paper describe potential risks incurred by study participants, whether such risks were disclosed to the subjects, and whether Institutional Review Board (IRB) approvals (or an equivalent approval/review based on the requirements of your country or institution) were obtained?
    \item[] Answer: \answerNA{} 
    \item[] Justification:
    \item[] Guidelines:
    \begin{itemize}
        \item The answer NA means that the paper does not involve crowdsourcing nor research with human subjects.
        \item Depending on the country in which research is conducted, IRB approval (or equivalent) may be required for any human subjects research. If you obtained IRB approval, you should clearly state this in the paper. 
        \item We recognize that the procedures for this may vary significantly between institutions and locations, and we expect authors to adhere to the NeurIPS Code of Ethics and the guidelines for their institution. 
        \item For initial submissions, do not include any information that would break anonymity (if applicable), such as the institution conducting the review.
    \end{itemize}

\end{enumerate}

\end{document}